\pdfoutput=1
\documentclass[11pt]{article}

\usepackage[]{acl}

\usepackage{times}
\usepackage{latexsym}

\usepackage[T1]{fontenc}
\usepackage[utf8]{inputenc}

\usepackage{microtype}

\usepackage{microtype}      % make text margins prettier
\usepackage{graphicx}       % import graphical images
\usepackage[T1]{fontenc}    % fix font related glitches such as "|"
\usepackage{multirow}       % merge rows in tables
\usepackage{subcaption}     % create sub-tables and sub-figures
\usepackage{amssymb}        % enable \mathbb, \mathcal
\usepackage{bold-extra}     % enable \texttt{\textbf{}}
\usepackage{bm}             % enable bold font in math mode
\usepackage[hang,flushmargin]{footmisc}  % minimize footnote indentation
\usepackage{amsmath}   % for matrices
\usepackage{makecell}  % for text spanning multiple lines in table cell
\usepackage{stfloats}  % for bottom-aligned table
\usepackage{arydshln}  % hdashline

\newcommand{\LN}{\linebreak\noindent}    % to manage inline spacing
\newcolumntype{?}{!{\vrule width 2pt}}

\usepackage{multicol} % added by gjw for formatting tab:novel-predicate-concept-accuracy

\usepackage{xcolor}
\definecolor{Blue}{HTML}{f0f8ff}
\definecolor{DarkBlue}{HTML}{003666}
\definecolor{Gray}{HTML}{54595f}
\usepackage{tcolorbox}
\tcbuselibrary{breakable}
\makeatletter
\def\tcb@finalize@environment{%
  \color{.}%
  \tcb@layer@dec%
}
\makeatother
\newtcolorbox{amr}[1]{colback=Blue,colframe=Blue,fontupper=\ttfamily,coltitle=Gray,title=\vspace*{.125cm}#1\vspace*{-.125cm}}

\usepackage{hyperref}
\definecolor{MedBlue}{HTML}{005299}
\hypersetup{colorlinks=true,linktocpage=true,allcolors=MedBlue}
\bibpunct{(}{)}{;}{a}{,}{,}

\usepackage{array}
\newcolumntype{x}[1]{>{\centering\let\newline\\\arraybackslash\hspace{0pt}}p{#1}} % width center cols
\newcommand{\see}[1]{\hyperref[#1]{\S \ref{#1}}} % hyperref command \see

\title{Widely Interpretable Semantic Representation: Frameless Meaning Representation for Broader Applicability}

\author{Lydia Feng \and Gregor Williamson \and Han He \and Jinho D. Choi \\
  Department of Computer Science \\
  Emory University \\
  Atlanta, GA, USA \\
  {\{lydia.feng, gregor.jude.williamson, han.he, jinho.choi\}@emory.edu}}
\begin{document}
\maketitle

\begin{abstract}
This paper presents a novel semantic representation, WISeR, that overcomes challenges for Abstract Meaning Representation (AMR).
Despite its strengths, AMR is not easily applied to languages or domains without predefined semantic frames, and its use of numbered arguments results in semantic role labels which are not directly interpretable and are semantically overloaded for parsers.
We examine the numbered arguments of predicates in AMR and convert them to thematic roles which do not require reference to semantic frames.
We create a new corpus of 1K English dialogue sentences annotated in both WISeR and AMR.
WISeR shows stronger inter-annotator agreement for beginner and experienced annotators, with beginners becoming proficient in WISeR annotation more quickly.
Finally, we train a state-of-the-art parser on the AMR 3.0 corpus and a WISeR corpus converted from AMR 3.0.
The parser is evaluated on these corpora and our dialogue corpus.
The WISeR model exhibits higher accuracy than its AMR counterpart across the board, demonstrating that WISeR is easier for parsers to learn.
\end{abstract}
\section{Introduction}
\label{sec:introduction}

Abstract Meaning Representations \citep[AMRs,][]{banarescu-etal-2013-abstract} represent the meaning of a natural language sentence as a singly rooted, directed acyclic graph in which nodes correspond to concepts and edges correspond to relations between them.
They are typically displayed using the human-readable PENMAN notation \citep{matthiessen1991text}, as shown in Figure~\ref{fig:amr}.
A central feature of AMR is its use of PropBank \citep{palmer-etal-2005-proposition, bonial-etal-2014-propbank}, a corpus of frames that assigns a specific argument structure to every predicate sense  in the form of a list of \textit{numbered arguments} (\texttt{:ARG}$n$). 

% Why AMR is good
There are several advantages of AMR, most notably its abstractness---meaning components need not be directly anchored to parts of the text (i.e., morphemes, lexical items, or MWEs).
In this respect, AMR differs from other graphical representation languages such as Elementary Dependency Structures \citep{oepen-lonning-2006-discriminant}, Prague Semantic Dependencies \cite{miyao-etal-2014-house}, and Universal Conceptual Cognitive Annotation \citep{abend-rappoport-2013-universal} which all feature some degree of anchoring \citep{oepen-etal-2019-mrp, oepen-etal-2020-mrp}.
Although this design feature has been argued to be a potential downside of AMR \citep{bender-etal-2015-layers}, it renders it particularly well-suited for parsing dialogue since it is able to represent the meaning of natural language expressions that are not syntactically well-formed, obviating problems that arise from many types of production error.
It also has a sizable corpus of annotation \cite{knight:2014a,knight:2017a,knight:2020a}, and a significant amount of research has been conducted to enhance AMR's representation of phenomena such as quantifier scope \citep{pustejovsky-etal-2019-modeling, lai-etal-2020-continuation, bos-2020-separating}, tense/aspect \citep{donatelli-etal-2018-annotation,donatelli2019tense}, intensional operators \citep{williamson-etal-2021-intensionalizing}, and speech acts \citep{bonial-etal-2020-dialogue}.
Lastly, numerous state-of-the-art AMR parsers have been developed with promising results \citep{cai-lam-2020-amr, oepen-etal-2020-mrp,xu-etal-2020-improving,lee-etal-2020-pushing,bevilacqua-etal-2021-one}.

Nonetheless, AMR has a few disadvantages.
It relies on PropBank for predicate argument structures, presupposing the existence of semantic frames for predicate senses. 
This makes it less versatile for languages or domains with frequent novel predicate senses, due to high upfront labor costs in creating many new frames.\footnote{A few studies have adapted AMR to other languages \cite{li-etal-2016-annotating,damonte-cohen-2018-cross, migueles-abraira-etal-2018-annotating, anchieta-pardo-2020-semantically,blloshmi-etal-2020-xl} and domains \cite{burns:16a}.}
Some studies that have adapted AMR to other languages have noted that AMR needs to be augmented to accommodate language-specific features.
For instance, \citet{migueles-abraira-etal-2018-annotating}
introduce novel predicates such as \texttt{sinnombre} ("nameless"), to annotate pro-drop in Spanish.

% Limitations of AMR parsing
Another problem for AMR is that numbered arguments are semantically opaque without reference to the frames.
There is no consistent mapping from numbered arguments to traditional thematic roles that is applicable to all senses besides perhaps \texttt{:ARG0} and \texttt{:ARG1}, which correspond to prototypical agent and patient respectively.
For instance, \texttt{:ARG2} of \texttt{tell-01} in Figure~\ref{fig:amr} is the entity the telling is directed at, while \texttt{:ARG2} of \texttt{dislodge-01} is the initial position of the dislodged entity.
Meanwhile, the initial position of the entity stepping-down is the \texttt{:ARG1} of \texttt{step-down-01}.
This inconsistent correspondence between numbered arguments and thematic roles makes semantic role labels uninterpretable for parsing models during training.

\begin{figure*}[htbp!]
\centering

\begin{subfigure}[t]{0.48\textwidth}
\resizebox{\columnwidth}{!}{
\texttt{\noindent
\begin{tabular}{@{}l@{}}
(t / tell-01 \\
\hphantom{(t }:ARG0 (w / woman) \\
\hphantom{(t }:ARG1 (s / step-down-04 \\
\hphantom{(t :ARG1 (s }:ARG0 w \\
\hphantom{(t :ARG1 (s }:ARG1 (r / role) \\
\hphantom{(t :ARG1 (s }:time (d / dislodge-01 \\
\hphantom{(t :ARG1 (s :time (d }:ARG0 w \\
\hphantom{(t :ARG1 (s :time (d }:ARG1 (b / boss) \\
\hphantom{(t :ARG1 (s :time (d }:ARG2 (b2 / board))) \\
\hphantom{(t }:ARG2 (m / man))
\end{tabular}
}}
\caption{AMR graph in Penman notation}
\label{fig:amr}
\end{subfigure}
~~~~~
\begin{subfigure}[t]{0.48\textwidth}
\resizebox{\columnwidth}{!}{
\texttt{\noindent
\begin{tabular}{@{}l@{}}
(t / tell \\
\hphantom{(t }:actor (w / woman) \\
\hphantom{(t }:theme (s / step-down \\
\hphantom{(t :theme (s }:actor w \\
\hphantom{(t :theme (s }:start (r / role) \\
\hphantom{(t :theme (s }:time (d / dislodge \\
\hphantom{(t :theme (s :time (d }:actor w \\
\hphantom{(t :theme (s :time (d }:theme (b / boss) \\
\hphantom{(t :theme (s :time (d }:start (b2 / board))) \\
\hphantom{(t }:benefactive (m / man))
\end{tabular}
}}
\caption{WISeR graph in Penman notation}
\label{fig:WISeR}
\end{subfigure}
\caption{AMR and WISeR graphs for the sentence \textit{`The woman told the man she will step down from the role when she dislodges the boss from the board'}.}
\label{fig:amr-and-wiser}
\vspace*{-0.5em}
\end{figure*}

% WISeR
Section~\ref{sec:drawbacks} discusses the drawbacks of AMR in detail.
Section~\ref{sec:WISeR} introduces a novel annotation scheme, WISeR (Widely Interpretable Semantic Representation), designed to overcome these challenges.
% Overcome limitations
In contrast to AMR, WISeR does not utilize frames, instead maintaining a one-to-one correspondence between argument labels and thematic roles.
It also has the benefit of permitting the introduction of novel predicates on an ad-hoc basis.
% IAA and speed
Section~\ref{sec:annotation-experiment} presents our new corpus comprising 1,000 English dialogue sentences annotated in both WISeR and AMR, making fair comparisons between the schemes for annotation adaptability and quality.
% Parsing
Section~\ref{sec:parsing-experiment} compares a seq-to-seq parsing model \citep{bevilacqua-etal-2021-one} trained on the AMR 3.0 corpus and a WISeR corpus converted from AMR 3.0.
Parsing models are evaluated on those corpora as well as our new dialogue corpora.\footnote{Our resources including the converted WISeR corpus and the new dialogue WISeR corpus are publicly available: \url{https://github.com/emorynlp/wiser}}

\section{Inside AMR}
\label{sec:drawbacks}

\subsection{Predicates in AMR}
\label{ssec:predicates-in-amr}

AMR annotation begins by identifying disambiguated predicate senses from PropBank frames.
Although providing frames as a reference to annotators is designed to ensure consistency during annotation, sense disambiguation is often difficult for annotators, leading to low agreement levels in word sense disambiguation tasks \cite{ng-etal-1999-case, lopez-de-lacalle-agirre-2015-crowdsourced}.
In addition, PropBank occasionally conflates senses (e.g., \texttt{put-on-08} is used for the sense in \textit{`put on some clothes'} as well as \textit{`put on a show'}).
AMR's reliance on PropBank means that it is constrained to only a few languages for which frames exist \cite{palmer-etal-2005-proposition,xue:05a,palmer:06a,zaghouani-etal-2010-revised,vaidya-etal-2011-analysis,duran-aluisio-2011-propbank,haverinen:15a,sahin:18a} and it often lacks domain-specific predicates in specialized domains.

However, AMR contains several predicate senses that are not found in PropBank.
These senses often represent idioms or multi-word constructions created ad-hoc during annotation (e.g., \texttt{throw-under-bus-08}, \texttt{pack-sand-00}).
Furthermore, there are 9 senses in AMR that have additional numbered arguments not featured in their respective PropBank frames.\footnote{The 9 senses with additional arguments in AMR:
\texttt{bind-01}: \texttt{ARG4}, \texttt{damage-01}: \texttt{ARG3}, \texttt{late-02}: \texttt{ARG3},
\texttt{misconduct-01}: \texttt{ARG1}, \texttt{oblige-02}: \texttt{ARG2}, \texttt{play-11}: \texttt{ARG3},
\texttt{raise-02}: \texttt{ARG3}, \texttt{rank-01}: \texttt{ARG5}, \texttt{unique-01}: \texttt{ARG3-4}.}
% \resizebox{0.97\columnwidth}{!}{\texttt{bind-01}: \texttt{ARG4}, \texttt{damage-01}: \texttt{ARG3}, \texttt{late-02}: \texttt{ARG3},}\\
% \resizebox{0.97\columnwidth}{!}{\texttt{misconduct-01}: \texttt{ARG1}, \texttt{oblige-02}: \texttt{ARG2}, \texttt{play-11}: \texttt{ARG3},}\\
% \resizebox{0.97\columnwidth}{!}{\texttt{raise-02}: \texttt{ARG3}, \texttt{rank-01}: \texttt{ARG5}, \texttt{unique-01}: \texttt{ARG3-4}}}

Table~\ref{tab:predicate-stats} shows the statistics of PropBank\footnote{English PropBank frames can be downloaded at \url{https://github.com/propbank/propbank-frames}} and the AMR 3.0 release \cite{knight:2020a}.
We calculate the number of frames in AMR 3.0 by combining information in the release text file\footnote{AMR frames are included in LDC2020T02 as\\\texttt{propbank-amr-frame-arg-descr.txt}} with the annotation corpus since there are frames in the text file that are not in the corpus, and vice versa.
Out of 9,090 senses in AMR 3.0, only 556 are unique to AMR.
In other words, 8,534 senses in AMR 3.0 (i.e., 94\%) are based on PropBank frames, emphasizing the extent to which AMR annotation depends on PropBank.
%We derive the frames of AMR 3.0 by combining information in the provided text file\footnote{AMR frames are included in LDC2020T02 as\\\texttt{propbank-amr-frame-arg-descr.txt}} with the actual annotation since frames in the text file do not form a superset of all the frames annotated in the corpus.\LN
%Out of 9,090 senses in AMR 3.0, only 556 frames are unique to AMR; in other words, 8,534 senses in AMR 3.0, that is 94\%, are borrowed from PropBank, indicating that AMR still hugely depends on the PropBank frames for its annotation.

\begin{table}[htbp!]
\centering\small
\begin{tabular}{l|r r}
\hline
 & \bf PropBank & \bf AMR 3.0 \\
\hline
Total \# of predicates  &  7,311 &  6,187 \\
Total \# of senses      & 10,687 &  9,090 \\
Total \# of arguments   & 27,012 & 23,171 \\
\# of unique predicates &  1,626 &    502 \\
\# of unique senses     &  2,153 &    556 \\
\hline
\end{tabular}
\caption{Statistics of PropBank and AMR 3.0.}
\label{tab:predicate-stats}
\end{table}

\subsection{Numbered Arguments in AMR}
\label{ssec:overloaded-num-args}

%The argument structures of predicate senses in PropBank consist of numbered arguments that do not directly encode meaning.
%As shown in Table~\ref{tab:arg-themroles}, the thematic roles \textit{benefactive} and \textit{attribute} may be encoded by either \texttt{ARG2} or \texttt{ARG3} in PropBank.
%As such, there is no one-to-one mapping between numbered arguments and thematic roles.
The argument structure of a predicate sense in PropBank is a set of numbered arguments.
As shown in Table~\ref{tab:arg-themroles}, the thematic role of benefactive or attribute may be encoded by either \texttt{:ARG2} or \texttt{:ARG3}.

\begin{table}[htbp!]
\centering\small
\begin{tabular}{c|l}
\hline
\textbf{Label} & \textbf{Thematic Role} \\
\hline
\texttt{ARG0} & agent \\ 
\texttt{ARG1} & patient \\
\texttt{ARG2} & instrument, benefactive, attribute \\ 
\texttt{ARG3} & starting point, benefactive, attribute \\
\texttt{ARG4} & ending point \\
\hline
\end{tabular}
\caption{Numbered arguments and corresponding thematic roles in the PB guidelines \cite{propbank}.}
\label{tab:arg-themroles}
\end{table}
\noindent
%\noindent \texttt{ARG0}/\texttt{ARG1} are intended to encode the thematic roles of prototypical agent/patient respectively.
%In practice, however, even this strong correspondence is occasionally lost.
%The semantics of numbered arguments can be accessed through two additional resources in PropBank: function tags and\LN VerbNet \cite{kipper:02a,Loper07combininglexical}.
Consequently, there is no one-to-one correspondence between numbered arguments and thematic roles.
For instance, \texttt{:ARG0}/\texttt{ARG1} largely correspond to the thematic roles of prototypical agent/patient respectively.
However, even this correspondence is occasionally lost.
As such, the numbered arguments do not directly encode meaning relations.
Rather, their semantics is given through two other resources in PropBank: function tags and VerbNet roles \cite{kipper:02a,kipper2005verbnet,Loper07combininglexical,kipper:08a}.
Table~\ref{tab:function-tags} shows the function tags used in PropBank.
%\footnote{The descriptions of these function tag acronyms are provided in Table~\ref{tab:function-tags} in Appendix~\ref{ssec:argument-descriptions}.}

\begin{table}[htbp!]
\centering\small
\begin{tabular}{ll|ll}
\hline
\textbf{Tag} & \textbf{Description} & \textbf{Tag} & \textbf{Description}\\
\hline
\texttt{PPT} & Prototypical Patient  & \texttt{EXT} & Extent\\
\texttt{PAG} & Prototypical Agent    & \texttt{CAU} & Cause\\
\texttt{GOL} & Goal                  & \texttt{COM} & Comitative\\
\texttt{PRD} & Secondary Predication & \texttt{PRP} & Purpose\\
\texttt{MNR} & Manner                & \texttt{TMP} & Temporal\\
\texttt{DIR} & Directional           & \texttt{ADJ} & Adjectival\\
\texttt{VSP} & Verb-specific         & \texttt{ADV} & Adverbial\\
\texttt{LOC} & Locative              & \texttt{REC} & Reciprocal\\\hline
\end{tabular}
\caption{Descriptions of the function tags in PropBank.}
\label{tab:function-tags}
\end{table}

Table~\ref{tab:ftag-argn} shows the distribution of function tags over numbered arguments, highlighting that every numbered argument is semantically opaque without reference to the PropBank frame.
As a result, numbered argument role labels make the task of automatic parsing more difficult for machines.

%While there was an effort to ensure that \texttt{ARG0} and \texttt{ARG1} corresponded to the traditional thematic roles of prototypical agent and prototypical patient/theme respectively, in practice even this correspondence is occasionally lost.
%We can access the semantics of numbered arguments through two additional resources in PropBank: function tags and VerbNet roles.
%The distribution of function tags over the numbered arguments in PropBank are summarized in Table~\ref{tab:ftag-argn} in Appendix \ref{sec:appendix}.
%This distribution highlights the fact that each numbered arguments is semantically overloaded, making the task of automatic classifiers more difficult.

%Note that the PropBank guidelines describe the thematic roles for \texttt{ARG0-4} (Table~\ref{tab:arg-themroles}).
%However, the latest PropBank frames define argument structures with \texttt{ARG5-6}, although they are rarely used.

\begin{table}[htbp!]
\centering\resizebox{\columnwidth}{!}{
\begin{tabular}{c|rrrrrrr|r}
%\multicolumn{10}{c}{\textbf{Numbered Arguments}} \\ 
\hline
 & \texttt{A0} & \texttt{A1} & \texttt{A2} & \texttt{A3} & \texttt{A4} & \texttt{A5} & \texttt{A6} & \hspace*{\fill}\bm{$\Sigma$}\hspace*{\fill} \\ 
\hline
%\multirow{16}{*}{\rotatebox{90}{\textbf{Function Tags}}} 
\texttt{PPT} &   389 & 8,593 & 1,249 &  49 &   4 &  0 & 0 & 10,284 \\  
\texttt{PAG} & 8,412 &   664 &    28 &   1 &   0 &  0 & 0 &  9,105 \\ 
\texttt{GOL} &     2 &   503 & 1,436 & 238 & 214 &  2 & 0 &  2,395 \\  
\texttt{PRD} &     0 &    79 &   701 & 231 &  85 & 10 & 0 &  1,106 \\  
\texttt{MNR} &     2 &    10 &   808 & 159 &   8 & 11 & 0 &    998 \\  
\texttt{DIR} &    18 &   147 &   518 & 270 &  14 &  4 & 0 &    971 \\  
\texttt{VSP} &     1 &    58 &   338 & 214 &  48 & 19 & 0 &    678 \\  
\texttt{LOC} &     6 &   196 &   268 &  43 &  25 &  4 & 0 &    542 \\  
\texttt{EXT} &     1 &     5 &   244 &  25 &   3 &  5 & 6 &    289 \\  
\texttt{CAU} &    75 &    22 &   140 &  30 &   0 &  0 & 0 &    267 \\  
\texttt{COM} &     0 &    83 &   100 &   9 &   4 &  0 & 0 &    196 \\  
\texttt{PRP} &     0 &     6 &    74 &  32 &   5 &  1 & 0 &    118 \\  
\texttt{TMP} &     0 &     3 &    15 &   3 &   6 &  1 & 0 &     28 \\  
\texttt{ADJ} &     0 &     5 &    10 &   4 &   0 &  0 & 0 &     19 \\  
\texttt{ADV} &     0 &     2 &     4 &   5 &   1 &  0 & 0 &     12 \\  
\texttt{REC} &     0 &     1 &     2 &   1 &   0 &  0 & 0 &      4 \\ 
\hline
\bm{$\Sigma$} & 8,906 & 10,377 & 5,935 & 1,314 & 417 & 57 & 6 & 27,012 \\ \hline
\end{tabular}}
\caption{Distribution of function tags (rows) over numbered arguments (columns) in PropBank.}
\label{tab:ftag-argn}
% \vspace{-1ex}
\end{table}

\noindent
As mentioned, numbered arguments are occasionally annotated with VerbNet roles.
Table~\ref{tab:vnrole-argn} shows the distribution of VerbNet thematic roles (in rows) over the numbered arguments (in columns) in PropBank frames.
Unfortunately, the coverage of PropBank frames associated with VerbNet classes is incomplete, with 25.5\% of PropBank frames not covered.
Even among the PropBank frames that are associated with VerbNet classes there are mismatches; an argument described in one resource may be omitted from the other, or a single argument may be split into multiple arguments.
These mismatches reflect both practical and theoretical differences in the resources, and as a result, only 40.6\% of arguments in PropBank are mapped to VerbNet roles. %\footnote{The distribution of VerbNet roles over numbered arguments is shown in Table~\ref{tab:vnrole-argn} in Appendix~\ref{ssec:argument-descriptions}.}
% Not all numbered arguments in the PropBank frames are aligned with VerbNet roles as only 40.6\% of arguments in these frames are mapped to specific VerbNet roles.

\begin{table*}[htbp!]
\centering\small
\begin{tabular}{l|rrrrrr|r}
\hline
& \texttt{ARG0} & \texttt{ARG1} & \texttt{ARG2} & \texttt{ARG3} & \texttt{ARG4} & \texttt{ARG5} & \multicolumn{1}{c}{\bm{$\Sigma$}} \\ 
\hline
 \texttt{agent} & 3,462 & 30 & 1 & 1 & 0 & 0 & 3,494 \\ 
 \texttt{theme} & 208 & 1,661 & 371 & 13 & 0 & 0 & 2,253 \\ 
 \texttt{patient} & 13 & 1,131 & 20 & 0 & 0 & 0 & 1,164 \\ 
 \texttt{experiencer} & 187 & 264 & 5 & 2 & 0 & 0 & 458 \\ 
 \texttt{destination} & 0 & 231 & 183 & 21 & 10 & 1 & 446 \\ 
 \texttt{stimulus} & 247 & 172 & 14 & 0 & 0 & 0 & 433 \\
 \texttt{location} & 7 & 145 & 142 & 30 & 23 & 1 & 348 \\ 
 \texttt{source} & 17 & 109 & 194 & 7 & 2 & 0 & 329 \\ 
 \texttt{recipient} & 0 & 56 & 251 & 10 & 0 & 0 & 317 \\ 
 \texttt{instrument} & 0 & 2 & 243 & 51 & 0 & 3 & 299 \\ 
 \texttt{topic} & 0 & 192 & 61 & 5 & 0 & 0 & 258 \\ 
 \texttt{co-patient} & 0 & 6 & 151 & 4 & 1 & 0 & 162 \\ 
 \texttt{beneficiary} & 0 & 40 & 47 & 44 & 7 & 0 & 138 \\ 
 \texttt{attribute} & 0 & 9 & 101 & 7 & 2 & 6 & 125 \\ 
 \texttt{result} & 0 & 30 & 81 & 5 & 7 & 0 & 123 \\ 
 \texttt{co-agent} & 0 & 69 & 25 & 0 & 0 & 0 & 94 \\ 
 \texttt{material} & 1 & 25 & 46 & 9 & 0 & 0 & 81 \\ 
 \texttt{goal} & 0 & 8 & 58 & 6 & 1 & 0 & 73 \\
 \texttt{co-theme} & 0 & 37 & 27 & 5 & 1 & 0 & 70 \\ 
 \texttt{product} & 0 & 35 & 17 & 4 & 13 & 0 & 69 \\ 
 \texttt{initial\_location} & 0 & 9 & 23 & 8 & 0 & 0 & 40 \\ 
 \texttt{cause} & 30 & 3 & 3 & 0 & 0 & 0 & 36 \\
 \texttt{asset} & 0 & 21 & 0 & 11 & 1 & 1 & 34 \\
 \texttt{predicate} & 0 & 4 & 18 & 6 & 0 & 0 & 28 \\
 \texttt{pivot} & 26 & 1 & 0 & 0 & 0 & 0 & 27 \\
 \texttt{extent} & 0 & 0 & 26 & 6 & 0 & 0 & 26 \\
 \texttt{value} & 0 & 5 & 13 & 7 & 0 & 0 & 25 \\
 \texttt{trajectory} & 0 & 3 & 0 & 0 & 0 & 0 & 3 \\ 
 \texttt{actor} & 1 & 0 & 0 & 0 & 0 & 0 & 1 \\ 
 \texttt{proposition} & 0 & 0 & 0 & 1 & 0 & 0 & 1 \\
\hline
 \multicolumn{1}{c|}{\bm{$\Sigma$}} & 4,199 & 4,298 & 2,121 & 257 & 68 & 12 & 10,955 \\ \hline
\end{tabular}
\caption{Distribution of VerbNet thematic roles over numbered arguments in PropBank.}
\label{tab:vnrole-argn}
\end{table*}

\section{Inside WISeR}
\label{sec:WISeR}

\subsection{Annotation Scheme}
\label{ssec:WISeR-scheme}

Here, we present the WISeR annotation scheme, which is designed to rectify the weaknesses of AMR in Section~\ref{sec:drawbacks}.
The complete WISeR annotation guidelines are provided in Appendix~\ref{sec:appendix}.
WISeR does not rely on frames, discarding numbered arguments and predicate sense IDs.
In this respect, WISeR annotation is similar to Open Information Extraction \citep[OpenIE,][]{banko2007,yates-etal-2007-textrunner, fader-etal-2011-identifying, angeli-etal-2015-leveraging}, the extraction of simple predications from large, diverse corpora without the need for a predefined vocabulary.
However, WISeR does make use of a set of thematic roles similar to VerbNet \cite{kipper2005verbnet} and meaning representations built on VerbNet, such as the Discourse Representation Structures used in the Parallel Meaning Bank \citep{abzianidze-etal-2017-parallel}.
WISeR represents thematic relations directly as edge labels, similar to the PENMAN Sentence Plan Language \cite{kasper-1989-flexible} and an earlier version of AMR prior to the incorporation of PropBank \cite{langkilde-knight-1998-generation}.
This particular design choice has strengths and weaknesses.
On the one hand, \citet{dowty1991thematic} famously argued that a closed set of discrete thematic roles is theoretically questionable, inspiring attempts to perform semantic role labelling without them \citep[e.g.,][]{reisinger-etal-2015-semantic, white-etal-2017-semantic}.
However, using thematic roles has practical benefits, allowing the categorization of arguments into natural classes and facilitating NLI about notions such as causation, direction of movement, secondary predication etc., as well as quantifying the frequency and distribution of thematic roles across different domains.

The WISeR graph in Figure~\ref{fig:WISeR} above shows how WISeR resolves the issues arising from use of numbered arguments in Figure~\ref{fig:amr}.
Both \textit{role} and \textit{board} stand in the \texttt{:start} relation to their predicates in WISeR because they both describe an initial state.
However, in AMR, the former is labeled \texttt{:ARG1} and the latter \texttt{:ARG2}.
Next, both \textit{man} and \textit{board} are labeled as \texttt{:ARG2} in AMR whereas they take distinct thematic roles of \texttt{:benefactive} and \texttt{:start} in WISeR.
Similarly, \texttt{:ARG1} is overloaded in AMR for \textit{role}, \textit{boss}, and \textit{man}, whereas WISeR disambiguates them by assigning the \texttt{:start} relation to \textit{role} and \texttt{:theme} to \textit{boss} and \textit{man}.

\begin{table*}[ht]
    \centering\small
    \begin{tabular}{ccccc}    \hline
    \textbf{ARGx} & \textbf{F-Tag} & \textbf{VerbNet Role} & \textbf{Description} & \textbf{WISeR Role} \\ \hline
    +ARG0 & +PAG & & & Actor  \\  
    +ARG0 & +CAU & & & Actor \\  
    +ARG1 & +PPT & & & Theme  \\  
    +ARG1 & +PAG & & +(entity$|$thing) & Theme  \\ 
     & +MNR & +instrument & & Instrument  \\ 
     & +MNR & -instrument & & Manner  \\ 
     & +GOL & +destination & & End \\
     & +GOL & & +\makecell{(end point$|$ending point$|$\\state$|$destination$|$attach$|$\\attached$|$target)} & End \\
     & +GOL & +\makecell{(beneficiary$|$recipient$|$\\experiencer)} & & Benefactive \\ 
     & +GOL & & \makecell{(benefactive$|$beneficiary$|$recipient$|$\\listener$|$hearer$|$perceiver$|$to whom$|$\\pay$|$paid)} & Benefactive \\ 
     & +LOC & +destination & & End \\ 
     & +LOC & +initial\_location & & Start \\ 
     & +LOC & +source & & Start \\
     & +LOC & -destination & & Location \\ 
     & +LOC & & \makecell{+(end point$|$ending point$|$state$|$\\destination$|$attach$|$target$|$end)} & End\\
     & +LOC & & +(start$|$source$|$from$|$starting) & Start \\
     & +DIR & +initial\_location & & Start  \\
     & +DIR & +source & & Start \\
     & +DIR & & +(start$|$source$|$from$|$starting) & Start \\ 
     & +COM & -recipient \& -beneficiary & & Accompanier \\ 
     & +COM & +(recipient$|$beneficiary) & & Benefactive \\
    +ARG1 & +VSP & +asset & & Theme \\ 
     & +VSP & & \makecell{+(price$|$money$|$rent$|$ \\ amount$|$gratuity)} & Asset \\ 
     & +PRP & & +(purpose$|$for) & Purpose \\ 
    -ARG1 & +CAU & -recipient & \makecell{+(why$|$reason$|$source$|$ \\ cause$|$crime$|$because)} & Cause \\ 
     & +VSP & +(material$|$source) & & Start \\
     & +VSP & & +(start$|$material$|$source) & Start \\ 
     & +VSP & & +(aspect$|$domain) \& -specific & Domain \\ 
    \hline
    \end{tabular}
    \caption{WISeR role mappings from ARGx, f-tag, VerbNet role, and description information.}
    \label{tab:conversion-rules}
\end{table*}

WISeR adopts non-core roles that exist in AMR, allowing annotation of most numbered arguments using these non-core roles. 
For example, WISeR incorporates the AMR \texttt{:source} role with numbered arguments corresponding to initial states into the role \texttt{:start}.
It also combines the \texttt{:beneficiary} role in AMR into the thematic role \texttt{:benefactive}.
This reduces redundancy in the annotation scheme since there are no longer two relations fulfilling the same semantic function.
WISeR also features a small number of thematic roles based on the PropBank function tags and VerbNet roles.
These include the \texttt{:actor} and \texttt{:theme} roles that broadly correspond to \texttt{:ARG0} and \texttt{:ARG1} in AMR, respectively.
The \texttt{:actor} role encompasses thematic agent as well as certain non-agentive subjects (e.g., \textit{the bus} in \textit{the bus hit the curb}).
% As a result, the inventory of WISeR roles is 2 fewer than the number of numbered arguments plus non-core roles in AMR 3.0.
Finally, WISeR adopts reified relations from AMR such as \texttt{have-rel-role} and \texttt{have-degree}.
The argument structure for each these reified relations is still semi-arbitrary and annotators will need to refer to the guidelines at first. %\footnote{The current annotation guidelines for WISeR can be found at our open-source project repository.}

\subsection{Converting AMR to WISeR}
\label{ssec:converting-amr}

To test the relative performance of parsing models on both AMR and WISeR, a mapping is defined to convert numbered arguments in the AMR 3.0 corpus into WISeR roles.
AMR 3.0 is the largest AMR corpus comprising 59,255 sentences collected from various sources including discussion forums, broadcast conversations, weblogs, newswire, children's stories, and more \cite{knight:2020a}.
There are 556 predicate senses in AMR 3.0 created on an ad-hoc basis (Section~\ref{ssec:predicates-in-amr}) without reference to a PropBank frame.
Sentences that include these ad-hoc senses are removed from this conversion.
Furthermore, sentences featuring rare predicates with highly-specific, non-generalizable argument structures are also removed.
For instance, \texttt{ARG1-9} of \texttt{publication-91} describe \textit{author}, \textit{title}, \textit{abstract}, \textit{text}, \textit{venue}, \textit{issue}, \textit{pages}, \textit{ID}, and \textit{editors}.
In total, there are 6 such predicates.\footnote{The 6 senses with non-generalizable argument structures are:
\texttt{byline-91}, \texttt{street-address-91}, \texttt{course-91}, \texttt{distribution-range-91}, \texttt{publication-91}, \texttt{sta}-\LN\texttt{tistical-test-91}. 
We hope to accommodate these predicates in future versions of WISeR.\label{fn:specialized-preds}}

\begin{table*}[ht!]
\centering\small
\begin{tabular}{l|rrrrrrr|r}
\hline
& \texttt{ARG0} & \texttt{ARG1} & \texttt{ARG2} & \texttt{ARG3} & \texttt{ARG4} & \texttt{ARG5} &\texttt{ARG6} & \multicolumn{1}{c}{\bm{$\Sigma$}} \\ 
\hline
\texttt{theme}       &    57 & 5,076 &   256 &  15 &   1 &  0 &  0 &  5,405 \\
\texttt{actor}       & 4,945 &    21 &     9 &   0 &   0 &  0 &  0 &  4,975 \\
\texttt{benefactive} &     1 &   148 &   554 &  90 &  38 &  2 &  0 &    833 \\
\texttt{end}         &     0 &   160 &   385 &  51 & 137 &  0 &  0 &    733 \\
\texttt{start}       &    14 &    63 &   322 & 190 &   6 &  0 &  0 &    595 \\
\texttt{instrument}  &     2 &     7 &   441 &  89 &   4 &  3 &  0 &    546 \\
\texttt{attribute}   &     0 &     6 &   144 &  44 &   6 &  2 &  0 &    202 \\
\texttt{location}    &     1 &    65 &    83 &   7 &   1 &  3 &  0 &    160 \\
\texttt{cause}       &     2 &    16 &   115 &  25 &   1 &  0 &  0 &    159 \\
\texttt{purpose}    &     0 &    11 &   122 &  19 &   5 &  1 &  0 &    158 \\
\texttt{topic}      &     2 &    14 &   113 &  20 &   3 &  0 &  0 &    152 \\
\texttt{accompanier} &     0 &    53 &    69 &   7 &   3 &  0 &  0 &    132 \\
\texttt{extent}      &     0 &     0 &    77 &   8 &   2 &  0 &  0 &     87 \\
\texttt{comparison}  &     0 &     1 &    51 &   7 &   3 &  3 &  2 &     67 \\
\texttt{asset}       &     0 &     1 &    11 &  53 &   1 &  0 &  0 &     66 \\
\texttt{domain}      &     0 &     4 &    23 &  11 &   0 &  0 &  0 &     38 \\
\texttt{mod}         &     0 &     2 &    15 &   4 &   1 &  0 &  0 &     22 \\
\texttt{manner}      &     0 &     3 &     9 &   5 &   2 &  0 &  0 &     19 \\
\texttt{direction}   &     0 &     0 &     7 &   0 &   2 &  5 &  0 &     14 \\
\texttt{path}        &     0 &     7 &     4 &   1 &   0 &  0 &  0 &     12 \\
\texttt{cause-of}    &     0 &     0 &     6 &   2 &   1 &  0 &  0 &      9 \\
\texttt{degree}      &     0 &     0 &     3 &   5 &   1 &  0 &  0 &      9 \\
\texttt{subevent}    &     0 &     0 &     3 &   2 &   1 &  0 &  0 &      6 \\
\texttt{quantity}    &     0 &     1 &     4 &   0 &   0 &  0 &  0 &      5 \\
\texttt{value}       &     0 &     0 &     3 &   2 &   0 &  0 &  0 &      5 \\
\texttt{time}        &     0 &     1 &     2 &   1 &   0 &  0 &  0 &      4 \\
\texttt{part-of}     &     0 &     1 &     1 &   2 &   0 &  0 &  0 &      4 \\
\texttt{duration}    &     0 &     0 &     2 &   0 &   1 &  0 &  0 &      3 \\
\texttt{theme-of}    &     0 &     0 &     2 &   0 &   0 &  0 &  0 &      2 \\
\texttt{range}       &     0 &     0 &     1 &   0 &   0 &  0 &  0 &      1 \\
\texttt{poss}        &     0 &     0 &     1 &   0 &   0 &  0 &  0 &      1 \\
\texttt{example}     &     0 &     0 &     0 &   1 &   0 &  0 &  0 &      1 \\
\texttt{consist-of}  &     0 &     0 &     1 &   0 &   0 &  0 &  0 &      1 \\
\texttt{concession}  &     0 &     0 &     1 &   0 &   0 &  0 &  0 &      1 \\
\texttt{frequency}&     0 &     0 &     0 &   1 &   0 &  0 &  0 &      1 \\
\hline
\multicolumn{1}{c|}{\bm{$\Sigma$}} & 5,024 & 5,661 & 2,840 & 662 & 220 & 19 &  2 & 14,428 \\ \hline
\end{tabular}
\caption{Distribution of PropBank numbered arguments to WISeR thematic roles.}
\label{tab:wiser-argn-ext}
\end{table*}

\noindent A total of 5,789 predicate senses are collected from PropBank frames that appear at least once in AMR 3.0.
The mapping converts every numbered argument for each of these senses to an appropriate WISeR role, totalling 15,120 unique arguments.
The conversion rules are presented in Table~\ref{tab:conversion-rules}. 
To define this mapping, several resources are used including the argument number, the function tag, the VerbNet role (if present), and certain keywords in the informal description of the argument written by PropBank annotators.
% These are used to convert numbered arguments into WISeR roles.
% Two or more of the following sources of information in PropBank are used to compute a conversion: the number of the argument, the functional tag, the VerbNet role (if present), and an informal description of the argument written by PropBank annotators.
For example, if an instance of an \texttt{ARG1} is labeled with a \texttt{PAG} function tag in PropBank and has a description containing either ``entity'' or ``thing'', then it is mapped to the WISeR role \texttt{theme} (see row 4 of Table~\ref{tab:conversion-rules}).
Using these mappings, for each AMR graph, all numbered argument edge labels were identified and relabeled with their WISeR role. 
We also relabeled AMR non-core roles of \texttt{source} to the WISeR role \texttt{start}, \texttt{destination} to \texttt{end}, \texttt{beneficiary} to \texttt{benefactive}, and \texttt{medium} to \texttt{manner}.
% Lastly, we converted concepts like \texttt{amr-unknown} and \texttt{amr-choice} into their WISeR counterparts.

The AMR-to-WISeR conversion rules result in a total of 12,311 mappings, which leaves 2,809 numbered arguments in AMR 3.0 that are not automatically mapped to WISeR roles.
%The automatic conversion results in a total of 12,311 mappings, leaving 2,809 numbered arguments which cannot be automatically mapped to WISeR roles.
These are manually mapped using the information in their PropBank frames as well as their specific usage in the corpus.
%While most arguments can be converted this way, there are 218 roles that are not used consistently in AMR 3.0.
%These are flagged for more investigation and examined to be mapped to WISeR roles.
Once all numbered arguments are converted into WISeR roles, sense IDs are removed so that the converted corpus becomes ``frameless''.

\noindent Table~\ref{tab:wiser-argn-ext} shows the distribution of numbered arguments over the 12 most frequently occurring roles in the converted WISeR corpus. % (the full version of this table is presented in Table~\ref{tab:wiser-argn-ext}).
Although mappings are created for 15,120 numbered arguments based on the PropBank frames, only 14,428 of them appear in the AMR 3.0 corpus, as shown in the $\Sigma$ column of the $\Sigma$ row in Table~\ref{tab:wiser-argn-ext}.

\section{WISeR Dialogue Corpus}
\label{sec:annotation-experiment}

This section presents our new WISeR corpus of 1,000 English language sentences from a variety of dialogue datasets such as EmpatheticDialogues \cite{DBLP:journals/corr/abs-1811-00207}, DailyDialog \cite{DBLP:journals/corr/abs-1710-03957}, Boston English Centre,\footnote{900 English Conversational Sentences from Boston English Centre: \url{https://youtu.be/JP5LYRTZtjw}} and PersonaChat \cite{gu2020dually}. 
Finally, we employ Mechanical Turking tasks to generate 300 sentences, in which subjects are provided with sentences from PersonaChat and asked to respond with emotionally driven reactions (100) or engaging follow-ups (200).\footnote{Crowd workers are compensated in-line with standard rates.}

500 of these sentences are split into 10 batches with every batch similar in length and complexity.
Six batches are split among beginner annotators and are double-annotated in both AMR and WISeR while the other four are divided evenly and double-annotated in either WISeR or AMR by experienced annotators.\footnote{The beginner annotators consist of 6 linguistics undergraduates who are rewarded for participation. The experienced annotators are the creators of the WISeR guidelines.}
All annotators annotate in both AMR and WISeR.
To control for familiarity, half of the annotators first annotate in AMR before switching to WISeR while the other half begin in WISeR before switching to AMR.

Beginner annotators are trained for a week and are given additional feedback on common errors, to minimize orthogonal differences in inter-annotator agreement.
The remaining 500 sentences are single-annotated by experienced annotators.
All annotation is performed using the StreamSide annotation tool \citep{choi2021streamside}.\footnote{\url{https://github.com/emorynlp/StreamSide}}

%\footnote{These included using the \texttt{poss} relation instead of a \texttt{have-rel-role(-91)} concept, the structure of agentive nouns, using \texttt{poss} instead of \texttt{part-of}, occasionally not using bare forms of verbs and nominals, misrepresenting named entities.
%We also advocated a consistent annotation for \textit{favorite} since this came up a lot in the dialogue and can reasonably be annotated in several different ways.}

\subsection{Inter-Annotator Agreement}
\label{ssec:inter-annotator-agreement}

Inter-annotator agreement (IAA) is measured via Smatch scores \citep{cai-knight-2013-smatch} on doubly-annotated batches. % that is the standard  evaluation metric to measure AMR parsing.
Table~\ref{tab:annoexp-smatch} shows IAA scores of individual batches and the macro-average scores of six batches by beginner and four batches by experienced annotators.
AMR and WISeR have similar IAA among experts; however, IAA for WISeR is noticeably higher among beginners, implying that AMR has a steeper learning curve, although both schemes produce high-quality annotation once annotators reach the expert-level.
All double-annotated sentences are adjudicated with correction prior to inclusion in the corpus.

\begin{table}[htbp!]
\centering\resizebox{0.9\columnwidth}{!}{
\begin{tabular}{c|cc|c|cc}
\hline
\multirow{2}{*}{\textbf{BID}} & \multicolumn{2}{c|}{\bf Beginner} & \multirow{2}{*}{\textbf{BID}} &\multicolumn{2}{c}{\bf Experienced} \\
 & \textbf{AMR} & \textbf{WISeR} &  & \textbf{AMR} & \textbf{WISeR}\\
\hline
01 & 0.72 & 0.74 & 07 & 0.87 & - \\
02 & 0.72 & 0.75 & 08 & 0.84 & - \\
03 & 0.68 & 0.70 & 09 & -    & 0.89 \\
04 & 0.69 & 0.79 & 10 & -    & 0.85 \\
05 & 0.77 & 0.79 & & & \\
06 & 0.72 & 0.76 & & & \\
\hline
\bm{$\mu_b$} & 0.72 & \bf 0.76 & \bm{$\mu_e$} & 0.86 & \bf 0.87 \\ \hline
\end{tabular}
}
\caption{IAA scores for batches annotated by beginner and expert annotators in AMR and WISeR. BID: batch ID, $\mu_{b/e}$: macro-average scores of the beginner and experienced groups, respectively.}
\label{tab:annoexp-smatch}
% \vspace*{-1.5ex}
\end{table}

\subsection{Annotation Time}
\label{ssec:annotation-time}

Every beginner annotator is assigned 3 batches and asked to report annotation times, allowing us to compare how quickly they become proficient in annotating in either scheme.
These results are summarized in Table~\ref{tab:annoexp-time}.
For Batches 1 and 2 there is practically no difference in time between AMR and WISeR annotation.
However, by Batch 3, annotating in WISeR is quicker.
This is likely due to familiarization with the WISeR guidelines and experience choosing the appropriate WISeR roles, while the process of identifying the correct frames and numbered arguments in AMR remains the same regardless of experience.

\begin{table}[htbp!]
\centering\resizebox{0.8\columnwidth}{!}{
\begin{tabular}{c|rrr|rrr}
\hline
\multirow{2}{*}{\textbf{AID}} & \multicolumn{3}{c|}{\bf AMR} & \multicolumn{3}{c}{\bf WISeR} \\
 & \multicolumn{1}{c}{\textbf{1}} & \multicolumn{1}{c}{\textbf{2}} & \multicolumn{1}{c|}{\textbf{3}} & \multicolumn{1}{c}{\textbf{1}} & \multicolumn{1}{c}{\textbf{2}} & \multicolumn{1}{c}{\textbf{3}} \\
\hline
A & 115 & 123 & 121 & 114 & 112 & 114 \\
B &  66 &  67 &  67 &  66 &  67 &  66 \\
C & 129 &  87 &  95 & 105 &  91 &  94 \\
D & 106 & 138 & 128 & 124 & 144 & 138 \\
E & 154 & 131 & 127 & 146 &  93 &  78 \\
F & 122 &  75 & \multicolumn{1}{c|}{-} & 140 & 105 & \multicolumn{1}{c}{-} \\
\hline
\bm{$\mu_a$} & \bf 115 & 104 & 107 & 116 & \bf 102 & \bf 98\\ \hline
\end{tabular}
}
\caption{Minutes per batch taken by the 6 annotators across 3 batches of 50 annotations. Annotator F completed only the first two batches. AID: annotator ID.}
\label{tab:annoexp-time}
% \vspace{-3ex}
\end{table}

%\begin{table*}[htbp!]
%\centering\small
%\begin{tabular}{l||r|r|r|r|r|r|r}
%\bf Source & \bf Sentences & \bf Tokens & \bf Concepts & \bf Relations & \bf Reentrancies & \bf Negations & \bf Names \\
%\hline\hline
%Boston English Center &   200 & 1,989   &  &  &    &     &  \\
%DailyDialog           &   200 & 2,177   &  &  &    &     &  \\
%EmpatheticDialogues   &   100 & 1,090   &  &  &    &     &  \\
%PersonaChat           &   200 & 1,431   &  &  &    &     &  \\
%MTurk-Reaction        &   100 & 298     &  &  &    &      &  \\
%MTurk-Followup        &   200 & 1,368   &  &  &    &      &  \\
%\hline\hline
%Total                 & 1,000 & 8,353   &  &  &    &    &  \\
%\end{tabular}
%\caption{Statistics of our AMR and converted WISeR dialogue corpus by different categories.}
%\label{tab:amr-wiser-corpus-stats}
%\end{table*}
% av token [9.945, 10.885, 10.9, 7.115, 2.98, 6.84]

\begin{table*}[htbp!]
\centering\resizebox{0.9\textwidth}{!}{
\begin{tabular}{c|rrr;{1.5pt/1.5pt}rr;{1.5pt/1.5pt}rr;{1.5pt/1.5pt}rr;{1.5pt/1.5pt}rr;{1.5pt/1.5pt}r}
\hline
\multirow{2}{*}{\bf Source} & \multirow{2}{*}{\bf Sent.} & \multirow{2}{*}{\bf Tokens} & \multicolumn{2}{c}{\bf Concepts} & \multicolumn{2}{c}{\bf Relations} & \multicolumn{2}{c}{\bf Reent.} & \multicolumn{2}{c}{\bf Negations} & \multicolumn{2}{c}{\bf NE} \\
  & & & \multicolumn{1}{c;{1.5pt/1.5pt}}{\bf A} & \multicolumn{1}{c}{\bf W} & \multicolumn{1}{c;{1.5pt/1.5pt}}{\bf A} & \multicolumn{1}{c}{\bf W} & \multicolumn{1}{c;{1.5pt/1.5pt}}{\bf A} & \multicolumn{1}{c}{\bf W} & \multicolumn{1}{c;{1.5pt/1.5pt}}{\bf A} & \multicolumn{1}{c}{\bf W} & \multicolumn{1}{c;{1.5pt/1.5pt}}{\bf A} & \multicolumn{1}{c}{\bf W} \\
\hline
\multicolumn{1}{l|}{DailyDialog}           &   200 & 2,177 & 1,297 & 1,298 & 1,315 & 1,318 & 211 & 229 &  27 &  26 & 21 & 22 \\
\multicolumn{1}{l|}{Boston English Center} &   200 & 1,989 & 1,182 & 1,196 & 1,167 & 1,179 & 217 & 219 &  33 &  33 & 12 & 13 \\
\multicolumn{1}{l|}{PersonaChat}           &   200 & 1,431 &   962 &   961 &   921 &   911 & 147 & 153 &  18 &  17 & 32 & 30 \\
\multicolumn{1}{l|}{EmpatheticDialogues}   &   100 & 1,090 &   692 &   699 &   712 &   710 & 131 & 128 &  20 &  20 &  1 &  1 \\
\multicolumn{1}{l|}{MTurk-Followup}        &   200 & 1,368 & 1,037 & 1,040 &   935 &   928 & 134 & 137 &   7 &   7 & 10 &  8 \\
\multicolumn{1}{l|}{MTurk-Reaction}        &   100 &   298 &   260 &   256 &   191 &   180 &  14 &  15 &   7 &   6 &  0 &  0 \\
\hline
\multicolumn{1}{c|}{\bm{$\Sigma$}}
                      & 1,000 & 8,353 & 5,433 & 5,447 & 5,240 & 5,226 & 854 & 881 & 112 & 109 & 76 & 74\\ \hline
\end{tabular}
}
\caption{Statistics of our dialogue corpus (in counts) by different categories annotated in AMR (A) and WISeR (W).\LN Sent: sentences, Reent: Reentrancies, NE: named entities.}
\label{tab:wiser-corpus-stats}
% \vspace*{-2ex}
\end{table*}

\subsection{Corpus Analytics}
\label{ssec:corpus-analytics}

Table~\ref{tab:wiser-corpus-stats} shows the statistics of our dialogue corpus annotated in AMR and WISeR, providing diverse utterances from six sources.
DailyDialog, Boston English Center, and EmpatheticDialogues have longer utterances as they are commonly in narrative form. 
PersonaChat consists of slightly shorter utterances, but its structures are still relatively complex.
Utterances in MTurk-Followup are mostly interrogatives and are shorter than ones from the other three.
MTurk-Reaction utterances are the shortest since they are mainly emotional reactions (e.g., \textit{that's impressive}).
These six sources yield 8.3K+ tokens with 5.4K+ concepts and 5.2K+ relations, allowing researchers to make meaningful parsing evaluation for dialogue.\footnote{At present, our corpus does not include \texttt{:wiki} information.
We intend to include this in a future release.}

%The dialogue corpus used is sourced from six different datasets to provide diverse utterances. 
%For example, DailyDialog, Boston English Center, and EmpatheticDialogues have more tokens per utterance as the utterance is commonly in narrative form. 
%This entails more concepts, relations, and reentrancies.
%On the other hand, PersonaChat statements are more declarative with less tokens, but the statements themselves are still relatively complex, with over 900 concepts, over 900 relations, and over 150 reentrancies.
%Sentences from PersonaChat often relate the speaker to a variety of other entities, contributing to the utterance's complexity. 
%MTurk-Followup displays a similar pattern of many concepts, relations, and reentrancies, as this dataset is mainly interrogatives. 
%Lastly, MTurk-Reaction displays the lowest token-per-sentence count, since this dataset is mainly short, emotional reactions such as "that's impressive" or "I understand". 
%These six sources combine to create our dialogue corpus of 1,000 sentences, over 8,000 tokens, and over 5,000 concepts and relations.

%No significant difference is found between AMR and WISeR in terms of statistics. %although WISeR generates slightly more concepts and reentrancies because it gives flexibility of creating new concepts as needed without defining new frames.
\noindent In comparison, the Dialogue-AMR corpus \cite{bonial-etal-2020-dialogue} consists of 80 hours of commands and requests made by humans to robots in search and navigation tasks.
It is mostly limited to these specific speech acts and mainly focuses on spatial words.
Our dialogue corpus, on the other hand, contains personal interactions about the speakers' likes and dislikes, relationships, and day-to-day life.
Finally, our corpus is publicly available whereas there is no public access currently available for the Dialogue-AMR corpus.

\section{Experiments}
\label{sec:parsing-experiment}

To assess the interpretability of the WISeR scheme, a parser is trained and tested on trimmed AMR 3.0 (AMR$_t$)\footnote{The AMR 3.0 corpus is trimmed as described in Section~\ref{ssec:converting-amr}.} and the WISeR corpus converted from AMR$_t$ (WISeR$_c$).
%To assess the effectiveness of our WISeR scheme, two state-of-the-art parsers (Sections \ref{ssec:cl20-parser} and \ref{ssec:b21-parser}) are trained on trimmed AMR 3.0 (AMR$_t$)\footnote{Sentences including ad-hoc predicates are removed in the trimmed AMR$_t$ corpus as described in Section~\ref{ssec:converting-amr}.} and the WISeR corpus converted from AMR3$_t$ (WISeR$_c$), and evaluated on AMR$_t$ and WISeR$_c$, respectively.
The AMR$_t$ parsing model is also tested on our dialogue corpus annotated in AMR (ADC).
Finally, the WISeR$_t$ model is evaluated on the ADC converted into WISeR (WDC$_c$), as well as our manually annotated WISeR dialogue corpus (WDC$_m$).
%Finally, the WISeR$_t$ models are evaluated on the WISeR corpus converted from ADC (WDC$_c$), which gives consistent annotation to WISeR$_c$, and our dialogue corpus manually annotated in WISeR (WDC$_m$).
The key differences between WDC$_c$ and WDC$_m$ are discussed in Section~\ref{ssec:parsing-limitations}.

\subsection{Datasets}

Table~\ref{tab:exp-datasets} shows the number of sentences in each split for the datasets used in our experiments.
ADC and WDC$_{c|m}$ are annotations of the same dialogue corpus and are used only for evaluation.

\begin{table}[htbp!]
\centering\resizebox{\columnwidth}{!}{
\begin{tabular}{c|ccc}
\hline
\bf Set & \bf AMR 3.0 & \bf AMR\bm{$_t$} | WISeR$_c$ & \bf ADC | WDC$_{c|m}$ \\
\hline
\tt TRN &      55,635 &      53,296 & - \\
\tt DEV & $\:\:$1,722 & $\:\:$1,656 & - \\
\tt TST & $\:\:$1,898 & $\:\:$1,813 & 1,000 \\
\hline
\bm{$\Sigma$} &      59,255 &      56,765 & 1,000 \\ \hline
\end{tabular}}
\caption{Number of sentences in the training (\texttt{TRN}), development (\texttt{DEV}), and evaluation (\texttt{TST}) sets.}
\label{tab:exp-datasets}
\end{table}

\subsection{Seq-to-Seq Parser}
\label{ssec:b21-parser}

We adopt a seq-to-seq parser, SPRING \cite{bevilacqua-etal-2021-one}, which holds the highest parsing accuracy on AMR 3.0 at the time of writing.
The hyper-parameter settings for the seq-to-seq parser (Section~\ref{ssec:b21-parser}) are described in Table~\ref{tbl:hyper-param-seq2seq}.

\begin{table}[htbp!]
	\centering
	\begin{tabular}{lr}
		\hline
		\multicolumn{2}{l}{\textbf{BART}} \\ \hline
		version & large \\
		\# parameters & 406M \\
		layers & 24 \\
		hidden size & 1024 \\
		heads & 16 \\
		\hline
		\multicolumn{2}{l}{\textbf{Adam Optimizer}} \\ \hline
		learning rate & 5e-5 \\
		warm up steps & 0 \\
        weight decay & 0.004 \\
        batch \#tokens & 5000 \\
        epochs & 30 \\
		\hline
	\end{tabular}
	\caption{Hyper-parameters for the seq-to-seq parser. }
	\label{tbl:hyper-param-seq2seq}
\end{table}

\noindent SPRING linearizes every graph into a sequence of tokens in the depth-first search order and trains the sequence using a seq-to-seq model, BART \cite{lewis-etal-2020-bart}.
In this sequence, special tokens are used to indicate variables and parentheses in the PENMAN notation.
Given a sentence and its linearized graph, BART is finetuned to learn the transduction from the former to the latter.
Once a linearized graph is generated, parenthesis parity is restored and any token that is not a possible continuation given the previous token is removed.
In our experiments, we use the BART large model with greedy decoding.

\begin{table*}[htbp!]
\centering
\resizebox{\textwidth}{!}{
\begin{tabular}{l|cccccccc} 
\hline
\textbf{Dataset} & \textbf{Smatch} & \textbf{Unlabeled} & \textbf{No WSD} & \textbf{Concepts} & \textbf{xSRL} & \textbf{Reentrancies} & \textbf{Negations} & \textbf{Named Entity} \\
\hline
AMR$_t$   &     83.5 $\pm$ 0.1 &     85.9 $\pm$ 0.0 &     84.0 $\pm$ 0.1 &   90.3 $\pm$ 0.0 & 75.9 $\pm$ 0.2 &    71.4 $\pm$ 0.3 &     73.0 $\pm$ 1.0 &     88.7 $\pm$ 0.5 \\ 
WISeR$_c$ & \bf 84.4 $\pm$ 0.1 & \bf 86.7 $\pm$ 0.1 & \bf 84.4 $\pm$ 0.1 & \bf 93.0 $\pm$ 0.1 & \bf 76.2 $\pm$ 0.4 & \bf 71.9 $\pm$ 0.2 & \bf 78.9 $\pm$ 0.2 & \bf 88.7 $\pm$ 0.4 \\
\hline
ADC       &     80.3 $\pm$ 0.2 &     83.8 $\pm$ 0.1 &     81.4 $\pm$ 0.2 &     86.8 $\pm$ 0.0 & 78.8 $\pm$ 0.3 &    71.8 $\pm$ 0.8 &     70.3 $\pm$ 0.5 &     65.5 $\pm$ 1.4 \\
WDC$_c$   & \bf 82.3 $\pm$ 0.2 &     85.7 $\pm$ 0.2 & \bf 82.3 $\pm$ 0.2 &     90.8 $\pm$ 0.1 & \bf 79.2 $\pm$ 0.3 & \bf 72.8 $\pm$ 0.3 &     76.2 $\pm$ 0.9 &     68.2 $\pm$ 1.8 \\
WDC$_m$   &     81.5 $\pm$ 0.2 & \bf 85.9 $\pm$ 0.2 &     81.5 $\pm$ 0.2 & \bf 91.1 $\pm$ 0.1 & 75.9 $\pm$ 0.2 &    70.6 $\pm$ 0.4 & \bf 78.2 $\pm$ 0.1 & \bf 74.9 $\pm$ 1.0 \\\hline
\end{tabular}}
\caption{Parsing performance achieved by the seq-to-seq model on the five evaluation sets over three runs.}
\label{tab:parse-results}
\vspace*{-1em}
\end{table*}

\subsection{Parsing Results}

%Table~\ref{tab:parse-results} shows the performance of the graph-based parser (GRB) and the seq-to-seq parser (S2S) on the five datasets.
Table~\ref{tab:parse-results} shows the performance of the seq-to-seq parser on the five datasets, with Smatch scores \cite{cai-knight-2013-smatch}, as well as more fine-grained metrics \cite{damonte-etal-2017-incremental}.
Comparing the results on AMR$_t$ and WISeR$_c$, the WISeR parser outperform the AMR parser on all categories, showing $\approx$1\% higher Smatch scores, which implies that WISeR is easier to learn, enabling parsers to train more robust models.
The \textit{No WSD} (no word sense disambiguation) scores for WISeR are equivalent to the Smatch scores because predicates in WISeR are not distinguished by senses.
Unsurprisingly, the WISeR parser shows higher scores on this category confirming that WSD introduces an extra burden on the AMR parser.
For \textit{Concepts} and \textit{Negations}, the WISeR parser also shows significant improvement over the AMR parser; $\approx$3\% and 6\%, respectively.

The \textit{SRL} (semantic role labeling) metric is only defined for numbered arguments and so is not applicable to WISeR.
To assess core argument labeling in both schemes, we propose a new metric called \textit{xSRL} (extended SRL).
The xSRL metric compares the WISeR roles in Table~\ref{tab:wiser-argn-ext} against \texttt{:ARG0-6} plus a few non-core roles in AMR, which correspond to the WISeR roles in Table~\ref{tab:wiser-argn-ext}.\footnote{\hbox{The non-core roles are: \texttt{:accompanier},}\\\hbox{\texttt{:beneficiary}, \texttt{:destination}, \texttt{:instrument},}\\\texttt{:location}, \texttt{:purpose}, \texttt{:source}, and \texttt{:topic}. \\The AMR role \texttt{:cause} is not used in the AMR 3.0 corpus.}
The WISeR parser again outperforms the AMR parser in this category.

\noindent Comparing the results on the ADC and WDC$_c$, which are out-of-domain datasets, we find the same trend.
The performance gain here is even larger as the WISeR parser produces a Smatch score higher by $\approx$2\%.
This indicates that the WISeR parser handles dialogue better.
Surprisingly, scores on the dialogue corpus are higher for \textit{xSRL} and \textit{Reentrancies} for both models.
This may be due to smaller graphs and possibly simpler argument structures in the dialogue corpus. %\footnote{Our experimental settings are provided in Appendix~\ref{ssec:experimental-settings}.}

Comparing the results of WDC$_c$ and WDC$_m$, we expect that WDC$_c$ should score better than WDC$_m$ due to discrepancies between converted and manual annotation.
However, the unlabeled scores are slightly higher on WDC$_m$ for both parsers, implying that the WISeR models still find the correct representations for out-of-domain data.
The named entity results of the seq-to-seq model are 6.5\% higher on WDC$_m$ than WDC$_c$ which is encouraging for areas such as Conversational AI that rely heavily on named entity recognition.

\subsection{Analysis}

%Concepts
We hypothesize that the seq-to-seq parser benefits from the more natural relation names in WISeR that are learnt during the pre-training of BART.
In addition, the WISeR parser has the freedom to coin novel concepts for predicate senses on which it lacks sufficient training.
For example, the verb \textit{premeditate} is absent from the training data, but present in the test set of AMR$_t$ and WISeR$_c$.
Out of 3 runs, the seq-to-seq AMR parser predicts the correct concept \texttt{premeditate-01} only once, predicting the concept \texttt{intend-01} once and \texttt{deliberate-01} once.
In comparison, the WISeR parser uses the novel concept \texttt{premeditate} every time.
The set of frames that occur only in the test set is rather small, so to make a fair comparison when evaluating the performance on the AMR$_{t}$ corpus, we restrict the comparison to the subset of novel frames that do not correspond to concepts in the WISeR$_{c}$ training data after conversion.\footnote{E.g., \texttt{move-04} is absent in the AMR training set but present in the test set. It is excluded from the comparison as it is converted to \texttt{move} which occurs in the WISeR training.}
When comparing on the dialogue corpus, we restrict our comparison to those concepts that are annotated identically in WDC$_{m}$ and WDC$_{c}$, and the concepts in AMR that feed into WDC$_{c}$.
We thus compare performance only on words that are translated into a novel predicate in every dataset.
The recall of the seq-to-seq parser across the evaluation sets is shown in Table~\ref{tab:novel-predicate-concept-accuracy}. 
We see that WISeR clearly outperforms AMR in recall of novel predicates.

\begin{table}[htbp!]
\centering\small
\begin{tabular}{lc|lc}
\hline
\textbf{Dataset} & \textbf{Recall} & \textbf{Dataset} & \textbf{Recall}\\
\hline
AMR$_{t}$   & 0.57  &  ADC       & 0.28 \\
WISeR$_{c}$ & 0.80  &  WDC$_{c}$ & 0.42 \\
            &       &  WDC$_{m}$ & 0.60\\
\hline
\end{tabular}

\caption{Recall of the seq-to-seq parser on novel predicate concepts in the five evaluation sets.}
\label{tab:novel-predicate-concept-accuracy}
\end{table}

%To get a clearer idea of how the parsers perform on edge-labelling, we computed a score for each edge label across all data sets. We then computed a weighted average for the 10 most frequent roles in each data set.

\noindent Finally, we tested the seq-to-seq parser on the WSD and SRL tasks independently.
The bottom left cell in Table~\ref{tab:WSD-vs-SRL} is the results for the WISeR parser, and the top right is the AMR parser.
The top left is a parser trained with PropBank senses and automatically converted WISeR roles, while the bottom right used numbered ARGs without predicate senses.\footnote{Since the choice of numbered argument depends on predicate sense IDs, WSD and SRL tasks are not sensibly separated with numbered arguments.}

\begin{table}[htbp!]
\centering\small

\begin{tabular}{c|cc}
\hline
 & \textbf{WISeR roles} & \textbf{Numbered ARGs} \\
\hline
+WSD & 83.8 $\pm$ 0.1 & 83.5 $\pm$ 0.1 \\
-WSD & 84.4 $\pm$ 0.1 & 84.2 $\pm$ 0.1 \\
\hline
\end{tabular}
\caption{Comparing the effect of transparent SRL and removing WSD independently.}
\label{tab:WSD-vs-SRL}
\end{table}

\noindent This shows a $\approx$0.3\% increase when using WISeR roles over numbered arguments even with predicate senses, while removing predicate senses accounts for a larger $\approx$0.7\% increase.

\subsection{Challenges}
\label{ssec:parsing-limitations}

A potential challenge in these experiments is that the converted WISeR corpus, WISeR$_c$, is arguably only pseudo-WISeR.
For instance, many predicate concepts corresponding to adjectives (e.g., \textit{great}) do not have PropBank frames.
Consequently, the sentence \textit{that is great} is annotated using the role \texttt{:domain} in AMR but \texttt{:theme} in WISeR.
Such inconsistency introduces noise to parsing models that leads to suboptimal performance.
We quantify the difference between the converted WISeR dialogue corpus (WDC$_{c}$) and the manually annotated corpus (WDC$_{m}$), by calculating their Smatch similarity, which returns a score of 0.88.
Although relatively high, this does indicate a training-evaluation discrepancy.
In future releases, we plan to enhance the automatic conversion to reduce this gap further.

\subsection{Discussions}
\label{ssec:discussion}

Since WISeR uses slightly fewer relations than AMR, we should perhaps expect the SRL classification task to be strictly simpler for WISeR.
However, this is not necessarily the case.
Table~\ref{tab:wiser-argn-ext} shows that the distribution of numbered arguments after \texttt{:ARG0-2} drop off rapidly (only 6\% of numbered arguments in AMR 3.0 are not \texttt{:ARG0-2}), whereas the distribution of WISeR roles shows a significantly shallower decline (22\% of core arguments are not covered by the 3 most frequent roles).
This larger number of reasonable candidate roles for a core argument in WISeR compared to AMR would ordinarily make classification harder.
A potential explanation for why the WISeR parser nonetheless outperforms the AMR parser is that many WISeR roles are associated with surface level syntax.
For example, a \texttt{:topic} argument is often introduced with the preposition \textit{about} or \textit{on}, an \texttt{:end} is typically introduced by \textit{to} etc.
These cues are obscured when a single numbered argument encodes more than one thematic role, or when one thematic role is encoded by more than one numbered argument.
In WISeR, there is a one-to-one correspondence between edge labels, and their semantic function.
As such, syntactic cues indicating the appropriate WISeR role can be identified, making classification easier.
Moreover, assigning consistent, more meaningful labels can help with data sparsity, while also capitalizing on the understanding that pre-trained models have of the language.

Finally, since automatically converted WISeR roles can be used with PropBank predicate senses, researchers can still make use of PropBank resources if they are required for downstream inference tasks, while nonetheless employing more transparent semantic role labels during parsing, albeit with more modest improvements.

\section{Conclusion}
\label{sec:conclusions}

% AMR relies on PropBank frames to disambiguate predicate senses and provide a predefined argument structure for each of these senses.
% This paper discusses several downsides of this approach.
% Due to the absence of appropriate frames, AMR is currently limited to a handful of languages.
% Also, numbered arguments in PropBank are semantically opaque, as each role (even \texttt{ARG0} and \texttt{ARG1}) encodes multiple thematic roles across frames.

To rectify a number of problems for AMR, this paper introduces a novel annotation scheme, WISeR, which allows for the spontaneous creation of predicates to extend to new domains and languages.
Our findings show that WISeR supports improved parsing performance as well as annotation of equal quality in less time.
We conclude that the removal of numbered arguments and sense disambiguation in favor of an open class of predicates and a modest inventory of thematic roles makes WISeR easier to learn for annotators and parsers alike.

% We will continue to explore new methods of improving WISeR and increase the size of our corpus in volume as well as diversity for other languages so that WISeR parsing models can be robust enough to be broadly used in practice.

% \input{tex/acknowledgements}

\bibliography{anthology,custom}
\bibliographystyle{acl_natbib}

\cleardoublepage
\appendix
\newpage\onecolumn
\section{WISeR Annotation Guidelines}
\label{sec:appendix}

WISeR is an annotation scheme that represents the semantics of an utterance. It attempts to revise the current AMR guidelines to be more interpretable by parsers. AMR makes use of PropBank frames which encode thematic roles using numbered arguments (\texttt{ARGx}).

\begin{table}[htbp!]
    \centering
    \begin{tabular}{| c || c |}
    \hline
        \texttt{ARG0} & agent \\ \hline
        \texttt{ARG1} & patient \\ \hline
        \texttt{ARG2} & instrument, benefactive, attribute \\ \hline
        \texttt{ARG3} & starting point, benefactive, attribute\\ \hline
        \texttt{ARG4} & ending point\\ \hline
        \texttt{ARGM} & modifier \\ \hline
    \end{tabular}
    \caption{List of arguments in PropBank}
    \label{tab:numbered-args}
\end{table}

Crucially, while thematic roles are meaning relations, numbered \texttt{ARG}s only indirectly encode meaning. In practice, there can be a disconnect between numbered \texttt{ARG}s and the thematic roles they are meant to encode. Notice that the benefactive and attribute roles can be encoded by both \texttt{ARG2} and \texttt{ARG3}. As a result, numbered \texttt{ARG}s are often semantically overloaded.

Additionally, many numbered \texttt{ARG}s are assigned the fine-grained role `verb specific' which does not form a meaningful class and thus may cause problems for parsers and annotators.

WISeR does not use numbered \texttt{ARG}s. Instead it encodes thematic role directly as WISeR relations. Below we give a comprehensive list of meaning relations available to WISeR 1.0. For each role, we will give some simple illustrative examples as well as some harder edge cases. We believe that by providing many different examples for each case, annotators will be able to use these guidelines as a reference manual when he or she is uncertain about choosing an appropriate representation. We have also included an index which includes traditional thematic roles and naive descriptions of relations to help annotators look up the appropriate WISeR relation. The relations are given roughly in order of frequency and are grouped together with other similar meaning relations.
WISeR adopts non-core roles from AMR and uses them in place of numbered \texttt{ARG2-6} and \texttt{ARGM}.
WISeR also introduces a small number of new roles based on VerbNet roles.

% \begin{table}[!htbp]
%     \centering
%     \begin{tabular}{| c || c |}
%     \hline
%         \textbf{AMR} & \textbf{WISeR} \\ \hline\hline
%         \texttt{ARG0} & \texttt{actor} \\ \hline
%         \texttt{ARG1} & \texttt{theme} \\ \hline
%         \texttt{ARG2} & \texttt{asset} \\ 
%         \texttt{ARG3} &  \texttt{attribute} \\ 
%         \texttt{ARG4} &  \texttt{comparison}\\ 
%         \texttt{ARG5} &  \\ 
%         \texttt{ARG6} & \\ 
%         \texttt{ARGM} & \\ \hline
%         \texttt{manner} & \texttt{manner} \\
%         \texttt{medium} & \\ \hline
%         other non-core roles & other non-core roles\\ \hline 
%     \end{tabular}
%     \caption{List of arguments in PropBank}
%     \label{tab:args-comparisons}
% \end{table}

% \vspace*{1cm}

\begin{tcolorbox}[colback=white,colframe=black,title=Note to annotators]

The WISeR annotation scheme shares many similarities with AMR. The primary differences are: \\ (i) disposing of sense relations, and (ii) replacing numbered \texttt{ARG}s with WISeR relations.

\medskip

If you encounter something not covered here, it will likely be covered in the AMR guidelines:

\url{https://github.com/amrisi/amr-guidelines/blob/master/amr.md}

\end{tcolorbox}

%%%%%%%%%%%%%%%%%%%%%%%%%%%%%%%%%%%%%%%%%%%%%%%%%%%%%%%%%%%%%%%%%%%%%%%%%%%%%%%%%%%%%%%%%%%%%%%%%%%%%%
%%%%%%%%%%%%%%%%%%%%%%%%%%%%%%%%%%%%%%%%%%%%%%%%%%%%%%%%%%%%%%%%%%%%%%%%%%%%%%%%%%%%%%%%%%%%%%%%%%%%%%
\subsection{Core roles}\label{core-roles}
%%%%%%%%%%%%%%%%%%%%%%%%%%%%%%%%%%%%%%%%%%%%%%%%%%%%%%%%%%%%%%%%%%%%%%%%%%%%%%%%%%%%%%%%%%%%%%%%%%%%%%
%%%%%%%%%%%%%%%%%%%%%%%%%%%%%%%%%%%%%%%%%%%%%%%%%%%%%%%%%%%%%%%%%%%%%%%%%%%%%%%%%%%%%%%%%%%%%%%%%%%%%%

%%%%%%%%%%%%%%%%%%%%%%%%%%%%%%%%%%%%%%%%%%%%%%%%%%%%%%%%%%%%%%%%%%%%%%%%%%%%%%%%%%%%%%%%%%%%%%%%%%%%%%
%%%%%%%%%%%%%%%%%%%%%%%%%%%%%%%%%%%%%%%%%%%%%%%%%%%%%%%%%%%%%%%%%%%%%%%%%%%%%%%%%%%%%%%%%%%%%%%%%%%%%%
\subsubsection{Actor}\label{actor}            %    Greg
%%%%%%%%%%%%%%%%%%%%%%%%%%%%%%%%%%%%%%%%%%%%%%%%%%%%%%%%%%%%%%%%%%%%%%%%%%%%%%%%%%%%%%%%%%%%%%%%%%%%%%
%%%%%%%%%%%%%%%%%%%%%%%%%%%%%%%%%%%%%%%%%%%%%%%%%%%%%%%%%%%%%%%%%%%%%%%%%%%%%%%%%%%%%%%%%%%%%%%%%%%%%%

WISeR makes use of the \texttt{actor} relation to encompass the traditional thematic role of agent.
\begin{amr}{The boy wants the girl to believe him}
(w / want \\ 
\hphantom{(w }:actor (b / boy) \\ 
\hphantom{(w }:theme (b2 / believe \\ 
\hphantom{(w :theme (b2 }:actor (g / girl) \\ 
\hphantom{(w :theme (b2 }:theme b))
\end{amr} \noindent
% This \hphantom command can be used to align the amr examples. \hphantom{blah} creates a space the length of the text `blah'
%
However, the \texttt{actor} relation is less specific than a thematic agent. An agent must be intentional, while the \texttt{actor} relation may also include non-intentional doers. The \texttt{actor} role corresponds to the thing which is the impetus behind the event.
\begin{amr}{The bus hit the curb}
(h / hit \\ 
\hphantom{(h }:actor (b / bus) \\ 
\hphantom{(h }:theme (c / curb))
\end{amr} \noindent
The role \texttt{actor} is also used to annotate the subject of a communication verb.
\begin{amr}{The boy said that the bus crashed}
(s / say \\ 
\hphantom{(s }:actor (b / boy) \\ 
\hphantom{(s }:theme (c / crash \\
\hphantom{(s :theme (c }:theme (b / bus)))
\end{amr} \noindent
Importantly, there is no one-to-one correspondence between the role of \texttt{actor} and the notion of grammatical  subject. Firstly, a subject is not always an actor (See also \texttt{theme} \see{theme} and \texttt{benefactive} \see{benefactive}).

Secondly, there are \texttt{actor} arguments which are not always grammatical subjects. For instance, WISeR (following PropBank) treats the entity or event which instils an emotion in a \texttt{theme} to be an \texttt{actor}.
\begin{amr}{The boy is scared of the monkey\\The monkey scares the boy}
(s / scare \\ 
\hphantom{(s }:actor (m / monkey) \\ 
\hphantom{(s }:theme (b / boy))
\end{amr} \noindent
Even when there is no transitive verbal form of the predicate (e.g., \textit{afraid}) the \texttt{actor} is still the entity which instils the emotion in the \texttt{theme}. In the following sentence, the monkey is the impetus of the fear.
\begin{amr}{The boy is afraid of the monkey}
(a / afraid \\ 
\hphantom{(a }:actor (m / monkey) \\ 
\hphantom{(a }:theme (b / boy))
\end{amr} \noindent
As mentioned, emotive predicates may even have an eventive \texttt{actor}. % Should we change this to cause?? Actor seems a bit silly
\begin{amr}{The boy is glad that the monkey left}
(g / glad \\ 
\hphantom{(g }:actor (l / leave \\
\hphantom{(g :actor (l }:actor (m / monkey)) \\
\hphantom{(g }:theme (b / boy))
\end{amr} \noindent
Finally, the subject of perception predicates (e.g., \textit{see} and \textit{hear}) is treated as an \texttt{actor} because it is doing the perceiving, even if unintentionally.\footnote{Those who are familiar with thematic roles might notice that we annotate thematic experiencers sometimes as a \texttt{theme} (as with emotive predicates like \textit{afraid}) and sometimes as an \texttt{actor} (as with verbs of perception like \textit{see}). Likewise, we sometimes annotate the so-called thematic stimulus as an \texttt{actor} (emotive predicates) and sometimes as a \texttt{theme} (verbs of perception). This is in keeping with PropBank, and we agree that it is the most natural way to annotate these constructions without introducing more relations.}
\begin{amr}{The boy saw the horse in the garden}
(s / see \\ 
\hphantom{(s }:actor (b / boy) \\
\hphantom{(s }:theme (h / horse \\
\hphantom{(s :theme (h }:location (g / garden)))
\end{amr} \noindent
%

%%%%%%%%%%%%%%%%%%%%%%%%%%%%%%%%%%%%%%%%%%%%%%%%%%%%%%%%%%%%%%%%%%%%%%%%%%%%%%%%%%%%%%%%%%%%%%%%%%%%%%
%%%%%%%%%%%%%%%%%%%%%%%%%%%%%%%%%%%%%%%%%%%%%%%%%%%%%%%%%%%%%%%%%%%%%%%%%%%%%%%%%%%%%%%%%%%%%%%%%%%%%%
\subsubsection{Theme}\label{theme}               %    Greg
%%%%%%%%%%%%%%%%%%%%%%%%%%%%%%%%%%%%%%%%%%%%%%%%%%%%%%%%%%%%%%%%%%%%%%%%%%%%%%%%%%%%%%%%%%%%%%%%%%%%%%
%%%%%%%%%%%%%%%%%%%%%%%%%%%%%%%%%%%%%%%%%%%%%%%%%%%%%%%%%%%%%%%%%%%%%%%%%%%%%%%%%%%%%%%%%%%%%%%%%%%%%%

WISeR does not distinguish between the thematic roles patient and theme. The role \texttt{theme} is used for arguments which either undergo an action or have some property.
\begin{amr}{The boy hugged the monkey}
(h / hug  \\ 
\hphantom{(b }:actor (b / boy)\\
\hphantom{(b }:theme (m / monkey))
\end{amr}\noindent
A \texttt{theme} may also appear as the grammatical subject. For instance, in an unaccusative construction.
\begin{amr}{The vase broke}
(b / break  \\ 
\hphantom{(b }:theme (v / vase))
\end{amr}\noindent
This retains its role in a causative construction when it occurs as the direct object and the \texttt{actor} is added as the grammatical subject.
\begin{amr}{The wind broke the vase}
(b / break  \\ 
\hphantom{(b }:actor (w / wind) \\ 
\hphantom{(b }:theme (v/ vase))
\end{amr} \noindent
A less obvious case of a \texttt{theme} is the subject of a verb like intransitive \textit{roll}.
\begin{amr}{The boy rolled down the hill}
(r / roll  \\ 
\hphantom{(r }:theme (b / boy) \\
\hphantom{(r }:direction (d / down) \\
\hphantom{(r }:path (h / hill))
\end{amr} \noindent
Compare this to the following.
\begin{amr}{The girl rolled the boy down the hill}
(r / roll  \\ 
\hphantom{(r }:actor (g / girl) \\
\hphantom{(r }:theme (b / boy) \\
\hphantom{(r }:direction (d / down) \\
\hphantom{(r }:path (h / hill))
\end{amr} \noindent
If it is clear from the context that the boy is the impetus behind the event, then the annotator can ascribe the concept \texttt{boy} both thematic relations.
\begin{amr}{The boy rolled down the hill on purpose}
(r / roll  \\ 
\hphantom{(r }:actor (b / boy) \\
\hphantom{(r }:theme b \\
\hphantom{(r }:direction (d / down) \\
\hphantom{(r }:path (h / hill) \\
\hphantom{(r }:manner (o / on-purpose))
\end{amr} \noindent
An even more striking example is the verb \textit{drive}.
\begin{amr}{The car drove west}
(d / drive  \\ 
\hphantom{(d }:theme (c / car) \\
\hphantom{(d }:direction (w / west))
\end{amr} \noindent
\begin{amr}{The girl drove the car west}
(d / drive  \\ 
\hphantom{(d }:actor (g / girl) \\
\hphantom{(d }:theme (c / car) \\
\hphantom{(d }:direction (w / west))
\end{amr} \noindent
As a rule of thumb, if the subject of an intransitive verb can also appear as the object when the verb is transitive, it is likely a \texttt{theme}.

The role \texttt{theme} is also used to annotate the message communicated by a communication verb.
\begin{amr}{The boy said that the bus crashed}
(s / say \\ 
\hphantom{(s }:actor (b / boy) \\ 
\hphantom{(s }:theme (c / crash \\
\hphantom{(s :theme (c }:theme (b / bus)))
\end{amr} \noindent
As well as propositions which are embedded under a modal concept.\footnote{More discussion of modality is included in the AMR guidelines.}
\begin{amr}{The boy can ski\\It is possible the boy is skiing\\The boy might ski}
(p / possible \\
\hphantom{(p }:theme (s / ski \\
\hphantom{(s :theme (g }:actor (b / boy)))
\end{amr} \noindent
\begin{amr}{The boy must clean the house\\The boy is obligated to clean the house\\It's obligatory that the boy clean the house}
(o / obligate \\
\hphantom{(o }:theme (c / clean \\
\hphantom{(o :theme (c }:actor (b / boy) \\
\hphantom{(o :theme (c }:theme (h / house)))
\end{amr} \noindent
The \texttt{theme} relation is also used when an argument has the property described by the predicate.
\begin{amr}{The girl is tall}
(t / tall  \\ 
\hphantom{(t }:theme (g / girl))
\end{amr} \noindent
\begin{amr}{The boy is glad}
(g / glad  \\ 
\hphantom{(t }:theme (b / boy))
\end{amr} \noindent
%

%%%%%%%%%%%%%%%%%%%%%%%%%%%%%%%%%%%%%%%%%%%%%%%%%%%%%%%%%%%%%%%%%%%%%%%%%%%%%%%%%%%%%%%%%%%%%%%%%%%%%%
%%%%%%%%%%%%%%%%%%%%%%%%%%%%%%%%%%%%%%%%%%%%%%%%%%%%%%%%%%%%%%%%%%%%%%%%%%%%%%%%%%%%%%%%%%%%%%%%%%%%%%
\subsubsection{Benefactive}\label{benefactive}            %    Greg
%%%%%%%%%%%%%%%%%%%%%%%%%%%%%%%%%%%%%%%%%%%%%%%%%%%%%%%%%%%%%%%%%%%%%%%%%%%%%%%%%%%%%%%%%%%%%%%%%%%%%%
%%%%%%%%%%%%%%%%%%%%%%%%%%%%%%%%%%%%%%%%%%%%%%%%%%%%%%%%%%%%%%%%%%%%%%%%%%%%%%%%%%%%%%%%%%%%%%%%%%%%%%

The \texttt{benefactive} role is used when representing a number of constructions. Most notably, it is used to represent a recipient in a dative or double object construction.
\begin{amr}{The girl gave a book to her friend\\The girl gave her friend a book}
(g / give \\
\hphantom{(g }:actor (g2 / girl) \\
\hphantom{(g }:theme (b / book) \\
\hphantom{(g }:benefactive (f / friend \\
\hphantom{(g :benefactive (f }:poss g2)))
\end{amr} \noindent
It is also used for some (but not all) other arguments introduced by prepositions such as \textit{to} and \textit{for} (See also \texttt{asset} \see{asset} and \texttt{purpose} \see{purpose}).
\begin{amr}{The girl sings to her cat\\The girl sings for her cat}
(s / sing \\
\hphantom{(s }:actor (g / girl) \\
\hphantom{(s }:benefactive (c / cat))
\end{amr} \noindent
The role \texttt{benefactive} is used when the argument is either a recipient or an individual/organisation for whose benefit or detriment an action is done (i.e., they are benefited or harmed by the event).
\begin{amr}{The dice fell kindly for the girl}
(f / fall \\
\hphantom{(f }:theme (d / dice) \\
\hphantom{(f }:manner (k / kind) \\
\hphantom{(f }:benefactive (g / girl))
\end{amr} \noindent
The role \texttt{benefactive} is also used to annotate the addressee or hearer of a communication verb.
\begin{amr}{The boy said to the girl that the bus crashed}
(s / say \\ 
\hphantom{(s }:actor (b / boy) \\ 
\hphantom{(s }:benefactive (g / girl) \\
\hphantom{(s }:theme (c / crash \\
\hphantom{(s :theme (c }:theme (b / bus)))
\end{amr} \noindent
\begin{amr}{The boy told the girl that it was raining}
(t / tell \\ 
\hphantom{(t }:actor (b / boy) \\ 
\hphantom{(t }:benefactive (g / girl) \\
\hphantom{(t }:theme (r / rain))
\end{amr} \noindent
\begin{amr}{The boy ordered the girl to clean her room}
(o / order \\ 
\hphantom{(o }:actor (b / boy) \\ 
\hphantom{(o }:benefactive (g / girl) \\
\hphantom{(o }:theme (c / clean \\
\hphantom{(o :theme (c }:actor g \\
\hphantom{(o :theme (c }:theme (r / room \\
\hphantom{(o :theme (c :theme (r }:poss g)))
\end{amr} \noindent
As well as arguments of permission and obligation modals.
\begin{amr}{The girl permitted the boy to eat a cookie}
(p / permit \\
\hphantom{(p }:actor (g / girl \\
\hphantom{(p }:benefactive (b / boy) \\
\hphantom{(p }:theme (e / eat \\
\hphantom{(p :theme (e }:actor b \\
\hphantom{(p :theme (e }:theme (c / cookie)))
\end{amr} \noindent
\begin{amr}{The girl obligated the boy to clean the house}
(o / obligate \\
\hphantom{(o }:actor (g / girl \\
\hphantom{(o }:benefactive (b / boy) \\
\hphantom{(o }:theme (c / clean \\
\hphantom{(o :theme (c }:actor b \\
\hphantom{(o :theme (c }:theme (h / house)))
\end{amr} \noindent
Notice that a \texttt{benefactive} role should not be confused with the notion of a beneficiary as the \texttt{benefactive} argument may be negatively affected.
\begin{amr}{The girl laid a trap for the monkey}
(l / lay \\
\hphantom{(l }:actor (g / girl) \\
\hphantom{(l }:theme (t / trap) \\
\hphantom{(l }:benefactive (m / monkey))
\end{amr} \noindent
Some verbs (e.g., \textit{receive}) seem like they should have a \texttt{benficative} subject. However, for the sake of maintaining consistency with both PropBank and VerbNet, we assign this subject the \texttt{actor} role.
\begin{amr}{The girl received a fine}
(r / receive \\
\hphantom{(r }:actor (g / girl) \\
\hphantom{(r }:theme (g2 / gift))
\end{amr} \noindent
%

%%%%%%%%%%%%%%%%%%%%%%%%%%%%%%%%%%%%%%%%%%%%%%%%%%%%%%%%%%%%%%%%%%%%%%%%%%%%%%%%%%%%%%%%%%%%%%%%%%%%%%
%%%%%%%%%%%%%%%%%%%%%%%%%%%%%%%%%%%%%%%%%%%%%%%%%%%%%%%%%%%%%%%%%%%%%%%%%%%%%%%%%%%%%%%%%%%%%%%%%%%%%%
\subsubsection{Asset}\label{asset}            %    Greg
%%%%%%%%%%%%%%%%%%%%%%%%%%%%%%%%%%%%%%%%%%%%%%%%%%%%%%%%%%%%%%%%%%%%%%%%%%%%%%%%%%%%%%%%%%%%%%%%%%%%%%
%%%%%%%%%%%%%%%%%%%%%%%%%%%%%%%%%%%%%%%%%%%%%%%%%%%%%%%%%%%%%%%%%%%%%%%%%%%%%%%%%%%%%%%%%%%%%%%%%%%%%%

In a bid to reduce verb specific arguments, WISeR makes use of the VerbNet relation \texttt{asset}. This is used with predicates which describe exchanges and transactions such as \texttt{buy}, \texttt{sell}, \texttt{offer}, \texttt{order}, as well as many others. We use the \texttt{asset} relation for any argument which moves in the opposite direction to the \texttt{theme} in an exchange.
\begin{amr}{The girl bought the axe for twenty dollars with a credit card\\The girl bought the axe with a credit card for twenty dollars}
(b / buy \\
\hphantom{(b }:actor (g / girl) \\
\hphantom{(b }:theme (a / axe) \\
\hphantom{(b }:asset (m / monetary-quantity \\
\hphantom{(b :asset (m }:quant \textcolor{DarkBlue}{20} \\
\hphantom{(b :asset (m }:unit (d / dollar))) \\
\hphantom{(b }:instrument (c / card \\
\hphantom{(b :instrument (c }:mod (c2 / credit)))
\end{amr} \noindent
Unlike the VerbNet role, the WISeR role \texttt{asset} is not restricted to monetary prices. For instance, verbs like \texttt{refund} or \texttt{rebate} have a \texttt{theme} which is typically a \texttt{monetary-quantity} while the \texttt{asset} is the thing exchanged in order to receive the money.
\begin{amr}{The cashier refunded the girl twenty dollars for the axe\\The cashier refunded twenty dollars to the girl for the axe}
(r / refund \\
\hphantom{(r }:actor (c / cashier) \\
\hphantom{(r }:benefactive (g / girl) \\
\hphantom{(r }:theme (m / monetary-quantity \\
\hphantom{(r :theme (m }:quant \textcolor{DarkBlue}{20} \\
\hphantom{(r :theme (m }:unit (d / dollar)) \\
\hphantom{(b }:asset (a / axe))
\end{amr} \noindent
%

%%%%%%%%%%%%%%%%%%%%%%%%%%%%%%%%%%%%%%%%%%%%%%%%%%%%%%%%%%%%%%%%%%%%%%%%%%%%%%%%%%%%%%%%%%%%%%%%%%%%%%
%%%%%%%%%%%%%%%%%%%%%%%%%%%%%%%%%%%%%%%%%%%%%%%%%%%%%%%%%%%%%%%%%%%%%%%%%%%%%%%%%%%%%%%%%%%%%%%%%%%%%%
\subsubsection{Instrument}\label{instrument}         %    Greg
%%%%%%%%%%%%%%%%%%%%%%%%%%%%%%%%%%%%%%%%%%%%%%%%%%%%%%%%%%%%%%%%%%%%%%%%%%%%%%%%%%%%%%%%%%%%%%%%%%%%%%
%%%%%%%%%%%%%%%%%%%%%%%%%%%%%%%%%%%%%%%%%%%%%%%%%%%%%%%%%%%%%%%%%%%%%%%%%%%%%%%%%%%%%%%%%%%%%%%%%%%%%%

The \texttt{instrument} relation is used for arguments which describe a thing used in carrying out an action. In English, \texttt{instruments} are usually introduced by the preposition \textit{with}.
\begin{amr}{The girl chopped the wood with the axe}
(c / chop \\
\hphantom{(c }:actor (g / girl) \\
\hphantom{(c }:theme (w / wood) \\
\hphantom{(c }:instrument (a / axe))
\end{amr} \noindent
Notice that, in the following example, \texttt{axe} is not an \texttt{instrument} of \texttt{chop} despite being used in the chopping event because it is not an argument of \texttt{chop}.
\begin{amr}{The girl used the axe to chop the wood}
(u / use \\
\hphantom{(u }:actor (g / girl) \\
\hphantom{(u }:theme (a / axe) \\
\hphantom{(u }:purpose (c / chop \\
\hphantom{(u :purpose (c }:actor g \\
\hphantom{(u :purpose (c }:theme (w / wood)))
\end{amr} \noindent
%

%%%%%%%%%%%%%%%%%%%%%%%%%%%%%%%%%%%%%%%%%%%%%%%%%%%%%%%%%%%%%%%%%%%%%%%%%%%%%%%%%%%%%%%%%%%%%%%%%%%%%%
%%%%%%%%%%%%%%%%%%%%%%%%%%%%%%%%%%%%%%%%%%%%%%%%%%%%%%%%%%%%%%%%%%%%%%%%%%%%%%%%%%%%%%%%%%%%%%%%%%%%%%
\subsubsection{Topic}\label{topic} % Lydia
%%%%%%%%%%%%%%%%%%%%%%%%%%%%%%%%%%%%%%%%%%%%%%%%%%%%%%%%%%%%%%%%%%%%%%%%%%%%%%%%%%%%%%%%%%%%%%%%%%%%%%
%%%%%%%%%%%%%%%%%%%%%%%%%%%%%%%%%%%%%%%%%%%%%%%%%%%%%%%%%%%%%%%%%%%%%%%%%%%%%%%%%%%%%%%%%%%%%%%%%%%%%%

The \texttt{topic} relation is used to annotate the subject-matter of an entity or event.
\begin{amr}{The professor wrote about math}
(w / write \\
\hphantom{(w }:actor (p / professor) \\
\hphantom{(w }:topic (m / math))
\end{amr} \noindent
\begin{amr}{The professor of math taught at the university\\The math professor taught at the university}
(t / teach \\
\hphantom{(t }:actor (p / professor \\
\hphantom{(t :actor (p }:topic (m / math)) \\
\hphantom{(t }:location (u / university))
\end{amr} \noindent
\begin{amr}{The employee emailed the president of human resources}
(e / email \\
\hphantom{(t }:actor (e2 / employee) \\
\hphantom{(t }:theme (p / president \\
\hphantom{(t :theme (p }:topic (r / resources \\
\hphantom{(t :theme (p :topic (r }:mod (h / human))))
\end{amr} \noindent
\begin{amr}{The problem with deregulation}
(p / problem \\
\hphantom{(p }:topic (d / deregulation))
\end{amr} \noindent
%

% ?? he is an expert [at fishing] ?? Topic or Attribute?

%%%%%%%%%%%%%%%%%%%%%%%%%%%%%%%%%%%%%%%%%%%%%%%%%%%%%%%%%%%%%%%%%%%%%%%%%%%%%%%%%%%%%%%%%%%%%%%%%%%%%%
%%%%%%%%%%%%%%%%%%%%%%%%%%%%%%%%%%%%%%%%%%%%%%%%%%%%%%%%%%%%%%%%%%%%%%%%%%%%%%%%%%%%%%%%%%%%%%%%%%%%%%
\subsubsection{Manner}\label{manner} % Lydia
%%%%%%%%%%%%%%%%%%%%%%%%%%%%%%%%%%%%%%%%%%%%%%%%%%%%%%%%%%%%%%%%%%%%%%%%%%%%%%%%%%%%%%%%%%%%%%%%%%%%%%
%%%%%%%%%%%%%%%%%%%%%%%%%%%%%%%%%%%%%%%%%%%%%%%%%%%%%%%%%%%%%%%%%%%%%%%%%%%%%%%%%%%%%%%%%%%%%%%%%%%%%%

The \texttt{manner} relation is used for arguments which describe the way something happens. It provides an answer to the question: ``how was it done?''. Notice that, when using \texttt{manner}, we drop the \textit{-ly} from the end of the adverb.
\begin{amr}{The boy sang beautifully}
(s / sing \\
\hphantom{(s }:actor (b / boy) \\
\hphantom{(s }:manner (b2 / beautiful))
\end{amr} \noindent
It can also represent the means by which something is done.
\begin{amr}{The boss decreased spending by shortening hours}
(d / decrease \\
\hphantom{(d }:actor (b / boss) \\
\hphantom{(d }:theme (s / spending) \\
\hphantom{(d }:manner (s2 / shorten \\
\hphantom{(d :manner (s2 }:actor b \\
\hphantom{(d :manner (s2 }:theme (h / hours)))
\end{amr} \noindent
WISeR also uses \texttt{manner} to represent arguments which AMR handles with a \texttt{medium} role. Note that this use of \texttt{manner} differs from \texttt{instrument}, as \texttt{instrument} relations describe the thing used, whereas this use of \texttt{manner} describes a more general means.
\begin{amr}{The girl talked in French}
(t / talk \\
\hphantom{(t }:actor (g / girl) \\
\hphantom{(t }:manner (l / language :wiki "French\_language" \\
\hphantom{(t :manner (l }:name (n / name :op1 "French")))
\end{amr} \noindent
\begin{amr}{The boy told the girl by email}
(t / tell \\
\hphantom{(t }:actor (b / boy) \\
\hphantom{(t }:theme (g / girl) \\
\hphantom{(t }:manner (e / email))
\end{amr} \noindent
%

%%%%%%%%%%%%%%%%%%%%%%%%%%%%%%%%%%%%%%%%%%%%%%%%%%%%%%%%%%%%%%%%%%%%%%%%%%%%%%%%%%%%%%%%%%%%%%%%%%%%%%
%%%%%%%%%%%%%%%%%%%%%%%%%%%%%%%%%%%%%%%%%%%%%%%%%%%%%%%%%%%%%%%%%%%%%%%%%%%%%%%%%%%%%%%%%%%%%%%%%%%%%%
\subsubsection{Accompanier}\label{accompanier} % Lydia
%%%%%%%%%%%%%%%%%%%%%%%%%%%%%%%%%%%%%%%%%%%%%%%%%%%%%%%%%%%%%%%%%%%%%%%%%%%%%%%%%%%%%%%%%%%%%%%%%%%%%%
%%%%%%%%%%%%%%%%%%%%%%%%%%%%%%%%%%%%%%%%%%%%%%%%%%%%%%%%%%%%%%%%%%%%%%%%%%%%%%%%%%%%%%%%%%%%%%%%%%%%%%

The \texttt{accompanier} relation is used for arguments that accompany another in the event. 
\begin{amr}{The nanny walked to town with the newborn}
(w / walk \\
\hphantom{(w }:actor (n / nanny) \\
\hphantom{(w }:end (t / town) \\
\hphantom{(w }:accompanier (n / newborn))
\end{amr} \noindent
This differs from an \texttt{actor} as the \texttt{accompanier} may not able to perform the event on its own.
\begin{amr}{The boy went to school with his backpack}
(g / go \\
\hphantom{(g }:actor (b / boy) \\
\hphantom{(g }:end (s / school) \\
\hphantom{(g }:accompanier (b2 / backpack \\
\hphantom{(g :accompanier (b2 }:poss b))
\end{amr} \noindent
%

%%%%%%%%%%%%%%%%%%%%%%%%%%%%%%%%%%%%%%%%%%%%%%%%%%%%%%%%%%%%%%%%%%%%%%%%%%%%%%%%%%%%%%%%%%%%%%%%%%%%%%
%%%%%%%%%%%%%%%%%%%%%%%%%%%%%%%%%%%%%%%%%%%%%%%%%%%%%%%%%%%%%%%%%%%%%%%%%%%%%%%%%%%%%%%%%%%%%%%%%%%%%%
\subsection{Spatial}\label{spatial}
%%%%%%%%%%%%%%%%%%%%%%%%%%%%%%%%%%%%%%%%%%%%%%%%%%%%%%%%%%%%%%%%%%%%%%%%%%%%%%%%%%%%%%%%%%%%%%%%%%%%%%
%%%%%%%%%%%%%%%%%%%%%%%%%%%%%%%%%%%%%%%%%%%%%%%%%%%%%%%%%%%%%%%%%%%%%%%%%%%%%%%%%%%%%%%%%%%%%%%%%%%%%%

%%%%%%%%%%%%%%%%%%%%%%%%%%%%%%%%%%%%%%%%%%%%%%%%%%%%%%%%%%%%%%%%%%%%%%%%%%%%%%%%%%%%%%%%%%%%%%%%%%%%%%
%%%%%%%%%%%%%%%%%%%%%%%%%%%%%%%%%%%%%%%%%%%%%%%%%%%%%%%%%%%%%%%%%%%%%%%%%%%%%%%%%%%%%%%%%%%%%%%%%%%%%%
\subsubsection{Location}\label{location} % GREG
%%%%%%%%%%%%%%%%%%%%%%%%%%%%%%%%%%%%%%%%%%%%%%%%%%%%%%%%%%%%%%%%%%%%%%%%%%%%%%%%%%%%%%%%%%%%%%%%%%%%%%
%%%%%%%%%%%%%%%%%%%%%%%%%%%%%%%%%%%%%%%%%%%%%%%%%%%%%%%%%%%%%%%%%%%%%%%%%%%%%%%%%%%%%%%%%%%%%%%%%%%%%%

The role \texttt{location} is used to represent constituents which describe where an event took place.
\begin{amr}{The man died in his house}
(d / die  \\ 
\hphantom{(m }:theme (m / man) \\
\hphantom{(m }:location (h / house \\
\hphantom{(m :location (h }:poss m))
\end{amr} \noindent
\begin{amr}{The man died near his house}
(d / die  \\ 
\hphantom{(m }:theme (m / man) \\
\hphantom{(m }:location (n / near \\
\hphantom{(m :location (n }:op1 (h / house \\
\hphantom{(m :location (n :op1 (h }:poss m)))
\end{amr} \noindent
\begin{amr}{The man died between his house and the river}
(d / die  \\ 
\hphantom{(m }:theme (m / man) \\
\hphantom{(m }:location (b / between \\
\hphantom{(m :location (n }:op1 (h / house \\
\hphantom{(m :location (n :op1 (h }:poss m) \\
\hphantom{(m :location (n }:op2 (r / river)))
\end{amr} \noindent
\begin{amr}{The detective arrived at the scene of the crime}
(a / arrive  \\ 
\hphantom{(a }:theme (d / detective) \\
\hphantom{(a }:end (s / scene \\
\hphantom{(a :end (s }:location-of (c / crime)))
\end{amr} \noindent
The \texttt{location} role can also be used for some verbal arguments.
\begin{amr}{The man fit three marshmallows in his mouth}
(f / fit  \\ 
\hphantom{(f }:actor (m / man) \\
\hphantom{(f }:theme (m2 / marshmallow \\
\hphantom{(f :theme (m2 }:quant \textcolor{DarkBlue}{3}) \\
\hphantom{(f }:location (m3 / mouth \\
\hphantom{(f :location (m3 }:part-of m))
\end{amr} \noindent
%

%%%%%%%%%%%%%%%%%%%%%%%%%%%%%%%%%%%%%%%%%%%%%%%%%%%%%%%%%%%%%%%%%%%%%%%%%%%%%%%%%%%%%%%%%%%%%%%%%%%%%%
%%%%%%%%%%%%%%%%%%%%%%%%%%%%%%%%%%%%%%%%%%%%%%%%%%%%%%%%%%%%%%%%%%%%%%%%%%%%%%%%%%%%%%%%%%%%%%%%%%%%%%
\subsubsection{Direction and Path}\label{direction and path}    %    Greg
%%%%%%%%%%%%%%%%%%%%%%%%%%%%%%%%%%%%%%%%%%%%%%%%%%%%%%%%%%%%%%%%%%%%%%%%%%%%%%%%%%%%%%%%%%%%%%%%%%%%%%
%%%%%%%%%%%%%%%%%%%%%%%%%%%%%%%%%%%%%%%%%%%%%%%%%%%%%%%%%%%%%%%%%%%%%%%%%%%%%%%%%%%%%%%%%%%%%%%%%%%%%%

The relations \texttt{direction} and \texttt{path} can represent arguments and modifiers of verbs of movement. Either one of these relations may be present without the other. In the following example we know the \texttt{path} of the bouncing, but not the direction.
\begin{amr}{The ball bounced along the street}
(b / bounce  \\ 
\hphantom{(b }:theme (b / ball) \\
\hphantom{(b }:path (a / along \\
\hphantom{(b :path (a }:op1 (s / street)))
\end{amr} \noindent
Similarly, we might know the \texttt{direction} but not the \texttt{path}.
\begin{amr}{The car drove west}
(d / drive  \\ 
\hphantom{(d }:theme (c / car) \\
\hphantom{(d }:direction (w / west))
\end{amr} \noindent
Besides the cardinal directions (\textit{north, south, east, west}), other typical directions include \textit{up, down, back, left, right, through, over}, etc.\footnote{For the sake of the annotation exercises we will not use the \texttt{wiki} role.}

A \texttt{direction} may also appear within a \texttt{path} argument.
\begin{amr}{The soldiers marched east along the road to Moscow}
(m / march  \\ 
\hphantom{(m }:actor (s / soldier) \\
\hphantom{(m }:direction (e / east) \\
\hphantom{(m }:path (a / along \\
\hphantom{(m :path (a }:op1 (r / road \\
\hphantom{(m :path (a :op1 (r }:direction (c / city \\
\hphantom{(m :path (a :op1 (r :direction (c }:wiki "Moscow" \\
\hphantom{(m :path (a :op1 (r :direction (c }:name (n / name \\
\hphantom{(m :path (a :op1 (r :direction (c :name (n }:op1 "Moscow")))))
\end{amr} \noindent
%

%%%%%%%%%%%%%%%%%%%%%%%%%%%%%%%%%%%%%%%%%%%%%%%%%%%%%%%%%%%%%%%%%%%%%%%%%%%%%%%%%%%%%%%%%%%%%%%%%%%%%%
%%%%%%%%%%%%%%%%%%%%%%%%%%%%%%%%%%%%%%%%%%%%%%%%%%%%%%%%%%%%%%%%%%%%%%%%%%%%%%%%%%%%%%%%%%%%%%%%%%%%%%
\subsubsection{Start and End}\label{start and end}         %    Greg
%%%%%%%%%%%%%%%%%%%%%%%%%%%%%%%%%%%%%%%%%%%%%%%%%%%%%%%%%%%%%%%%%%%%%%%%%%%%%%%%%%%%%%%%%%%%%%%%%%%%%%
%%%%%%%%%%%%%%%%%%%%%%%%%%%%%%%%%%%%%%%%%%%%%%%%%%%%%%%%%%%%%%%%%%%%%%%%%%%%%%%%%%%%%%%%%%%%%%%%%%%%%%

The relations \texttt{start} and \texttt{end} are generally used for changes in location (corresponding to the AMR roles \texttt{source} and \texttt{destination}) or changes in state. Examples of locative \texttt{start} and \texttt{end} are given below.
\begin{amr}{The monkey jumped from tree to tree}
(j / jump \\
\hphantom{(j }:actor (m / monkey) \\
\hphantom{(j }:start (t / tree) \\
\hphantom{(j }:end (t2 / tree))
\end{amr} \noindent
\begin{amr}{{He drove west, from Houston to Austin}}
(d / drive \\
\hphantom{(d }:actor (h / he) \\
\hphantom{(d }:direction (w / west) \\
\hphantom{(d }:start (c / city :wiki ``Houston'' \\
\hphantom{(d :start (c }:name (n / name :op1 ``Houston'')) \\
\hphantom{(d }:end (c2 / city :wiki ``Austin,\_Texas'' \\
\hphantom{(d :end (c2 }:name (n2 / name :op1 "Austin")))
\end{amr} \noindent
They can also be used for more abstract directional arguments.
\begin{amr}{They are descended from royalty}
(d / descend \\
\hphantom{(d }:theme (t / they) \\
\hphantom{(d }:start (r / royalty))
\end{amr} \noindent
WISeR also uses \texttt{start} for initial states or materials in verbs of creation (i.e., the \texttt{material} role of VerbNet), and \texttt{end} for the thing created (i.e., the \texttt{product} role of VerbNet).
\begin{amr}{She cast the bronze into a statue\\She cast a statue out of bronze}
(c / cast \\
\hphantom{(c }:actor (h / he) \\
\hphantom{(c }:start (b / bronze) \\
\hphantom{(c }:end (s / statue))
\end{amr} \noindent
\begin{amr}{She made a dress out of her curtains\\She made her curtains into a dress}
(m / make \\
\hphantom{(m }:actor (s / she) \\
\hphantom{(m }:start (c / curtains \\
\hphantom{(m ::start (c }:poss s)) \\
\hphantom{(m }:end (d / dress))
\end{amr} \noindent
\begin{amr}{He folded the paper into a card\\He folded a card out of the paper}
(f / fold \\
\hphantom{(f }:actor (s / she) \\
\hphantom{(f }:start (p / paper) \\
\hphantom{(f }:end (c / card))
\end{amr} \noindent
As well as certain verb specific arguments.
\begin{amr}{The monkey arranged the bananas from a neat stack into a messy pile}
(a / arrange \\
\hphantom{(a }:actor (m / monkey) \\
\hphantom{(a }:theme (b / banana) \\
\hphantom{(a }:start (s / stack \\
\hphantom{(a :start (s }:mod (n / neat) \\
\hphantom{(a }:end (p / pile \\
\hphantom{(a :end (p }:mod (m2 / messy))
\end{amr} \noindent
Annotators should also be careful with locative alternations. These involve a \texttt{theme} and an \texttt{end}.
\begin{amr}{He sprayed paint onto the wall\\He sprayed the wall with paint}
(s / spray \\
\hphantom{(s }:theme (p / paint) \\
\hphantom{(s }:end (w / wall))
\end{amr} \noindent
\begin{amr}{He loaded hay onto the cart\\He loaded the cart with hay}
(l / load \\
\hphantom{(l }:theme (h / hay) \\
\hphantom{(l }:end (c / cart))
\end{amr} \noindent
However, the \texttt{end} can also appear without the \texttt{theme}. So annotators should be particularly careful not to assign the \texttt{theme} role to the \texttt{end} here.
\begin{amr}{He sprayed the wall}
(s / spray \\
\hphantom{(s }:end (w / wall))
\end{amr} \noindent
\begin{amr}{He loaded the cart}
(l / load \\
\hphantom{(l }:end (c / cart))
\end{amr} \noindent
An annotator might also wonder whether we could also annotate a \texttt{benefactive} as an \texttt{end} in a transfer of possession verbs. For instance, in the following example. 
\begin{amr}{The girl gave a dog to the boy}
(g / give \\
\hphantom{(g }:actor (g / girl) \\
\hphantom{(g }:theme (d / dog) \\
\hphantom{(g }:benefactive (b / boy))
\end{amr} \noindent
WISeR opts to prioritize \texttt{benefactive} above \texttt{end}. If the argument could best be described as a ``recipient'', as in this example, you should use \texttt{benefactive}.

Finally, it might be hard to tell the difference between an \texttt{end} and a \texttt{direction}.
\begin{amr}{The girl threw the pie at the boy}
(t / throw \\
\hphantom{(b }:actor (g / girl) \\
\hphantom{(b }:theme (p / pie) \\
\hphantom{(b }:end (b / boy))
\end{amr} \noindent
Typically, a \texttt{direction} is a word such as \textit{up, down, left, right, north, south, east, west, over, under, through} etc. or a place like a country or city (See \texttt{direction} \see{direction and path}).

%%%%%%%%%%%%%%%%%%%%%%%%%%%%%%%%%%%%%%%%%%%%%%%%%%%%%%%%%%%%%%%%%%%%%%%%%%%%%%%%%%%%%%%%%%%%%%%%%%%%%%
%%%%%%%%%%%%%%%%%%%%%%%%%%%%%%%%%%%%%%%%%%%%%%%%%%%%%%%%%%%%%%%%%%%%%%%%%%%%%%%%%%%%%%%%%%%%%%%%%%%%%%
\subsection{Temporal}\label{temporal}
%%%%%%%%%%%%%%%%%%%%%%%%%%%%%%%%%%%%%%%%%%%%%%%%%%%%%%%%%%%%%%%%%%%%%%%%%%%%%%%%%%%%%%%%%%%%%%%%%%%%%%
%%%%%%%%%%%%%%%%%%%%%%%%%%%%%%%%%%%%%%%%%%%%%%%%%%%%%%%%%%%%%%%%%%%%%%%%%%%%%%%%%%%%%%%%%%%%%%%%%%%%%%

%%%%%%%%%%%%%%%%%%%%%%%%%%%%%%%%%%%%%%%%%%%%%%%%%%%%%%%%%%%%%%%%%%%%%%%%%%%%%%%%%%%%%%%%%%%%%%%%%%%%%%
%%%%%%%%%%%%%%%%%%%%%%%%%%%%%%%%%%%%%%%%%%%%%%%%%%%%%%%%%%%%%%%%%%%%%%%%%%%%%%%%%%%%%%%%%%%%%%%%%%%%%%
\subsubsection{Time}\label{time} % Lydia 
%%%%%%%%%%%%%%%%%%%%%%%%%%%%%%%%%%%%%%%%%%%%%%%%%%%%%%%%%%%%%%%%%%%%%%%%%%%%%%%%%%%%%%%%%%%%%%%%%%%%%%
%%%%%%%%%%%%%%%%%%%%%%%%%%%%%%%%%%%%%%%%%%%%%%%%%%%%%%%%%%%%%%%%%%%%%%%%%%%%%%%%%%%%%%%%%%%%%%%%%%%%%%
The \texttt{time} relation establishes when an event took place.
\begin{amr}{The robbery happened yesterday}
(r / robbery  \\ 
\hphantom{(f }:time (y / yesterday)
\end{amr} \noindent
\begin{amr}{The bridge was built in December}
(b / build  \\ 
\hphantom{(b }:theme (b2 / bridge) \\
\hphantom{(b }:time (d / date-entity \\
\hphantom{(b :time (d }:month \textcolor{DarkBlue}{12}))
\end{amr} \noindent
It can also be used for relative time.
\begin{amr}{The woman had just eaten lunch}
(e / eat  \\ 
\hphantom{(e }:actor (w / woman) \\
\hphantom{(e }:theme (l / lunch) \\
\hphantom{(e }:time (r / recent))
\end{amr} \noindent
In addition, the \texttt{time} relation can equate the time of two events.
\begin{amr}{The woman frowned when the baby cried}
(f / frown  \\ 
\hphantom{(f }:actor (w / woman) \\
\hphantom{(f }:time (c / cry) \\
\hphantom{(f :time (c }:actor (b / baby)))
\end{amr} \noindent
%

%%%%%%%%%%%%%%%%%%%%%%%%%%%%%%%%%%%%%%%%%%%%%%%%%%%%%%%%%%%%%%%%%%%%%%%%%%%%%%%%%%%%%%%%%%%%%%%%%%%%%%
%%%%%%%%%%%%%%%%%%%%%%%%%%%%%%%%%%%%%%%%%%%%%%%%%%%%%%%%%%%%%%%%%%%%%%%%%%%%%%%%%%%%%%%%%%%%%%%%%%%%%%
\subsubsection{Duration}\label{duration} % Lydia
%%%%%%%%%%%%%%%%%%%%%%%%%%%%%%%%%%%%%%%%%%%%%%%%%%%%%%%%%%%%%%%%%%%%%%%%%%%%%%%%%%%%%%%%%%%%%%%%%%%%%%
%%%%%%%%%%%%%%%%%%%%%%%%%%%%%%%%%%%%%%%%%%%%%%%%%%%%%%%%%%%%%%%%%%%%%%%%%%%%%%%%%%%%%%%%%%%%%%%%%%%%%%

The \texttt{duration} relation describes the amount of time over which an event occurs.
\begin{amr}{He worked for two hours yesterday}
(w / work  \\ 
\hphantom{(w }:actor (h / he) \\
\hphantom{(w }:duration (t / temporal-quantity \\
\hphantom{(w :duration (t }:quant \textcolor{DarkBlue}{2} \\
\hphantom{(w :duration (t }:unit (h2 / hour)) \\
\hphantom{(w }:time (y / yesterday)) 
\end{amr} \noindent
\begin{amr}{The investigator searched for a long time}
(s / search  \\ 
\hphantom{(s }:actor (i / investigator) \\
\hphantom{(s }:duration (l / long)
\end{amr} \noindent
\begin{amr}{The athlete finished the marathon in two hours}
(f / finish  \\ 
\hphantom{(f }:actor (a / athlete) \\
\hphantom{(f }:theme (r / run) \\
\hphantom{(f :theme (r }:actor a \\
\hphantom{(f :theme (r }:theme (m / marathon) \\
\hphantom{(f :theme (r }:duration (t / temporal-quantity \\
\hphantom{(f :theme (r :duration (t }:quant \textcolor{DarkBlue}{2} \\
\hphantom{(f :theme (r :duration (t }:unit (h / hour))) 
\end{amr} \noindent
%

%%%%%%%%%%%%%%%%%%%%%%%%%%%%%%%%%%%%%%%%%%%%%%%%%%%%%%%%%%%%%%%%%%%%%%%%%%%%%%%%%%%%%%%%%%%%%%%%%%%%%%
%%%%%%%%%%%%%%%%%%%%%%%%%%%%%%%%%%%%%%%%%%%%%%%%%%%%%%%%%%%%%%%%%%%%%%%%%%%%%%%%%%%%%%%%%%%%%%%%%%%%%%
\subsubsection{Frequency}\label{frequency}   % Lydia
%%%%%%%%%%%%%%%%%%%%%%%%%%%%%%%%%%%%%%%%%%%%%%%%%%%%%%%%%%%%%%%%%%%%%%%%%%%%%%%%%%%%%%%%%%%%%%%%%%%%%%
%%%%%%%%%%%%%%%%%%%%%%%%%%%%%%%%%%%%%%%%%%%%%%%%%%%%%%%%%%%%%%%%%%%%%%%%%%%%%%%%%%%%%%%%%%%%%%%%%%%%%%
The \texttt{frequency} relation describes how often something occurs.
\begin{amr}{The phone rang three times}
(r / ring  \\ 
\hphantom{(r }:theme (p / phone) \\
\hphantom{(r }:frequency \textcolor{DarkBlue}{3})
\end{amr} \noindent
It can also be used to represent quantificational temporal adverbs.
\begin{amr}{She always eats breakfast}
(e / eat  \\ 
\hphantom{(e }:actor (s / she) \\
\hphantom{(e }:theme (b / breakfast) \\ 
\hphantom{(e }:frequency (a / always))
\end{amr} \noindent
%

%%%%%%%%%%%%%%%%%%%%%%%%%%%%%%%%%%%%%%%%%%%%%%%%%%%%%%%%%%%%%%%%%%%%%%%%%%%%%%%%%%%%%%%%%%%%%%%%%%%%%%
%%%%%%%%%%%%%%%%%%%%%%%%%%%%%%%%%%%%%%%%%%%%%%%%%%%%%%%%%%%%%%%%%%%%%%%%%%%%%%%%%%%%%%%%%%%%%%%%%%%%%%
\subsubsection{Range}\label{range}    % Lydia
%%%%%%%%%%%%%%%%%%%%%%%%%%%%%%%%%%%%%%%%%%%%%%%%%%%%%%%%%%%%%%%%%%%%%%%%%%%%%%%%%%%%%%%%%%%%%%%%%%%%%%
%%%%%%%%%%%%%%%%%%%%%%%%%%%%%%%%%%%%%%%%%%%%%%%%%%%%%%%%%%%%%%%%%%%%%%%%%%%%%%%%%%%%%%%%%%%%%%%%%%%%%%
%
The \texttt{range} relation is used to describe a period of time over which an event occurs. This is different from \texttt{duration}, because it does not measure the length of the event. Rather, it establishes a period of time in which the event occurs.
\begin{amr}{His first drink in 3 years}
(d / drink  \\ 
\hphantom{(d }:actor (h / he) \\
\hphantom{(d }:ord (o / ordinal-entity \\
\hphantom{(d :ord (o }:value \textcolor{DarkBlue}{1} \\
\hphantom{(d :ord (o }:range (t / temporal-quantity \\
\hphantom{(d :ord (o :range (t }: quant \textcolor{DarkBlue}{3} \\
\hphantom{(d :ord (o :range (t }: unit (y / year))))
\end{amr} \noindent
Notice in the next example that if we had used the \texttt{duration} role, the sentence would mean \textit{``it did not snow for 10 years''}, which is compatible with it having snowed for 9 years.
\begin{amr}{It had not snowed in ten years}
(s / snow  \\ 
\hphantom{(s }:polarity - \\
\hphantom{(s }:range (t / temporal-quantity \\
\hphantom{(s :range (t }:quant \textcolor{DarkBlue}{10} \\
\hphantom{(s :range (t }:unit (y / year)))
\end{amr} \noindent
%

%%%%%%%%%%%%%%%%%%%%%%%%%%%%%%%%%%%%%%%%%%%%%%%%%%%%%%%%%%%%%%%%%%%%%%%%%%%%%%%%%%%%%%%%%%%%%%%%%%%%%%
%%%%%%%%%%%%%%%%%%%%%%%%%%%%%%%%%%%%%%%%%%%%%%%%%%%%%%%%%%%%%%%%%%%%%%%%%%%%%%%%%%%%%%%%%%%%%%%%%%%%%%
\subsection{Causal/Conditional/Concessive}\label{causal/concessive}
%%%%%%%%%%%%%%%%%%%%%%%%%%%%%%%%%%%%%%%%%%%%%%%%%%%%%%%%%%%%%%%%%%%%%%%%%%%%%%%%%%%%%%%%%%%%%%%%%%%%%%
%%%%%%%%%%%%%%%%%%%%%%%%%%%%%%%%%%%%%%%%%%%%%%%%%%%%%%%%%%%%%%%%%%%%%%%%%%%%%%%%%%%%%%%%%%%%%%%%%%%%%%

%%%%%%%%%%%%%%%%%%%%%%%%%%%%%%%%%%%%%%%%%%%%%%%%%%%%%%%%%%%%%%%%%%%%%%%%%%%%%%%%%%%%%%%%%%%%%%%%%%%%%%
%%%%%%%%%%%%%%%%%%%%%%%%%%%%%%%%%%%%%%%%%%%%%%%%%%%%%%%%%%%%%%%%%%%%%%%%%%%%%%%%%%%%%%%%%%%%%%%%%%%%%%
\subsubsection{Cause}\label{cause} % GREG
%%%%%%%%%%%%%%%%%%%%%%%%%%%%%%%%%%%%%%%%%%%%%%%%%%%%%%%%%%%%%%%%%%%%%%%%%%%%%%%%%%%%%%%%%%%%%%%%%%%%%%
%%%%%%%%%%%%%%%%%%%%%%%%%%%%%%%%%%%%%%%%%%%%%%%%%%%%%%%%%%%%%%%%%%%%%%%%%%%%%%%%%%%%%%%%%%%%%%%%%%%%%%

The \texttt{cause} role is typically used for causal adverbial clauses such as \textit{because} clauses. The \texttt{cause} role is used to annotate an answer to the question ``why did the event happen?''.
\begin{amr}{The wind broke the vase because it was fragile}
(b / break \\
\hphantom{(b }:actor (w / wind) \\
\hphantom{(b }:theme (v / vase) \\
\hphantom{(b }:cause (f / fragile) \\
\hphantom{(b :cause (f }:theme v))
\end{amr} \noindent
A \texttt{cause} is one of two ways of representing the notion of a ``reason'' in WISeR, (See also \texttt{purpose} \see{purpose}). In the following sentence, there are two reasons the judge sentenced the man.
\begin{amr}{The judge sentenced the man for speeding because he looked shifty}
(s / sentence \\
\hphantom{(s }:actor (j / judge) \\
\hphantom{(s }:theme (m / man) \\
\hphantom{(s }:cause (a / and \\
\hphantom{(b :cause (a }:op1 (s2 / speed \\
\hphantom{(b :cause (a :op1 (s2 }:actor m) \\
\hphantom{(b :cause (a }:op2 (s3 / seem \\
\hphantom{(b :cause (a :op2 (s3 }:theme (s4 / shifty \\
\hphantom{(b :cause (a :op2 (s3 :theme (s4 }:theme m))))
\end{amr} \noindent
WISeR also uses the inverse \texttt{cause-of} relation to represent some result states.\footnote{Notice we use \texttt{cause-of} instead of \texttt{end} here, since \textit{pieces} is not something which is made out of the vase (i.e. a product). Moreover, it is not a grammatical argument of \textit{break}. Finally, it does not take part in the material/product alternation which is indicative of the \texttt{end} relation.\\ i.\,\,\,He folded the paper into a card / He folded a card out of the paper
\\ii.\,\,He broke the vase into pieces / *He broke pieces out of the vase}
\begin{amr}{The vase broke into pieces}
(b / break \\
\hphantom{(b }:theme (v / vase) \\
\hphantom{(b }:cause-of (i / in-pieces \\
\hphantom{(b :cause-of (i }:theme v))
\end{amr} \noindent
\begin{amr}{He painted the house green}
(p / paint \\
\hphantom{(p }:actor (h / he) \\
\hphantom{(p }:theme (h2 / house) \\
\hphantom{(s }:cause-of (g / green \\
\hphantom{(s :cause-of (d }:theme h))
\end{amr} \noindent
\begin{amr}{The soldiers marched themselves tired}
(m / march \\
\hphantom{(m }:actor (s / soldier) \\
\hphantom{(m }:cause-of (t / tired \\
\hphantom{(m :cause-of (t }:theme s))
\end{amr} \noindent
%

%%%%%%%%%%%%%%%%%%%%%%%%%%%%%%%%%%%%%%%%%%%%%%%%%%%%%%%%%%%%%%%%%%%%%%%%%%%%%%%%%%%%%%%%%%%%%%%%%%%%%%
%%%%%%%%%%%%%%%%%%%%%%%%%%%%%%%%%%%%%%%%%%%%%%%%%%%%%%%%%%%%%%%%%%%%%%%%%%%%%%%%%%%%%%%%%%%%%%%%%%%%%%
\subsubsection{Purpose}\label{purpose} % GREG
%%%%%%%%%%%%%%%%%%%%%%%%%%%%%%%%%%%%%%%%%%%%%%%%%%%%%%%%%%%%%%%%%%%%%%%%%%%%%%%%%%%%%%%%%%%%%%%%%%%%%%
%%%%%%%%%%%%%%%%%%%%%%%%%%%%%%%%%%%%%%%%%%%%%%%%%%%%%%%%%%%%%%%%%%%%%%%%%%%%%%%%%%%%%%%%%%%%%%%%%%%%%%

The role \texttt{purpose} is used to annotate an answer to the question ``why was the event done?''. A \texttt{purpose} is one of two ways to represent a ``reason'' in WISeR, (See also \texttt{cause} \see{cause}). In contrast to a \texttt{cause}, a \texttt{purpose} always follows the event.
\begin{amr}{She works for a living}
(w / work \\
\hphantom{(m }:actor (s / she) \\
\hphantom{(m }:purpose (l / living))
\end{amr} \noindent
\begin{amr}{She works to improve her life}
(w / work \\
\hphantom{(m }:actor (s / she) \\
\hphantom{(m }:purpose (i / improve\\
\hphantom{(m :purpose (i }:theme (l / life \\
\hphantom{(m :purpose (i :theme (l }:poss h)))
\end{amr} \noindent
A physical object may also have a \texttt{purpose}.
\begin{amr}{She found a trap for catching monkeys}
(f / find \\
\hphantom{(f }:actor (s / she) \\
\hphantom{(f }:theme (t / trap \\
\hphantom{(f :theme (t }:purpose (c / catch \\
\hphantom{(f :theme (t :purpose (c }:theme (m / monkey))))
\end{amr} \noindent
%

%%%%%%%%%%%%%%%%%%%%%%%%%%%%%%%%%%%%%%%%%%%%%%%%%%%%%%%%%%%%%%%%%%%%%%%%%%%%%%%%%%%%%%%%%%%%%%%%%%%%%%
%%%%%%%%%%%%%%%%%%%%%%%%%%%%%%%%%%%%%%%%%%%%%%%%%%%%%%%%%%%%%%%%%%%%%%%%%%%%%%%%%%%%%%%%%%%%%%%%%%%%%%
\subsubsection{Condition}\label{condition}   % GREG
%%%%%%%%%%%%%%%%%%%%%%%%%%%%%%%%%%%%%%%%%%%%%%%%%%%%%%%%%%%%%%%%%%%%%%%%%%%%%%%%%%%%%%%%%%%%%%%%%%%%%%
%%%%%%%%%%%%%%%%%%%%%%%%%%%%%%%%%%%%%%%%%%%%%%%%%%%%%%%%%%%%%%%%%%%%%%%%%%%%%%%%%%%%%%%%%%%%%%%%%%%%%%

The \texttt{condition} role is used for introducing an \textit{if}-clause.
\begin{amr}{We will stay home if it rains}
(s / stay \\
\hphantom{(s }:theme (w / we) \\
\hphantom{(s }:location (h / home) \\
\hphantom{(s }:condition (r / rain))
\end{amr} \noindent
In combination with \texttt{polarity}, it can be used to represent an \textit{unless} clause.\footnote{The AMR guidelines incorrectly places the negative polarity directly under the root concept, rather than embedded within the \texttt{condition}. Our example shows that this is incorrect. Consider the following sentences.\smallskip
\\ \hspace*{.5cm} i.\,\,\,\,We will win the tournament unless we lose the final game.
\\ \hspace*{.5cm} ii.\,\,\,We won't win the tournament if we lose the final game.
\\ \hspace*{.5cm} iii.\,\,We will win the tournament if we don't lose the final game.\smallskip\\
Both (ii) and (iii) would be true if (i) is true. However, (ii) would be true even if we cannot win the tournament with a draw. But (i) and (iii) would be false. This shows that (i) is closer in meaning to (iii) than (ii).}
\begin{amr}{We will win the tournament unless we lose the final game}
(w / win \\
\hphantom{(w }:actor (w2 / we) \\
\hphantom{(w }:theme (t / tournament) \\
\hphantom{(w }:condition (l / lose \\
\hphantom{(w :condition (l }:polarity - \\
\hphantom{(w :condition (l }:actor (w2 / we) \\
\hphantom{(w :condition (l }:theme (g / game \\
\hphantom{(w :condition (l :theme (g }:mod (f / final))))
\end{amr} \noindent
It can also represent an unconditional \textit{whether or not} clause.
\begin{amr}{We will go to the park whether it rains or not}
(g / go \\
\hphantom{(g }:actor (w / we) \\
\hphantom{(g }:end (p / park) \\
\hphantom{(g }:condition (o / or \\
\hphantom{(g :condition (o }:op1 (r / rain) \\
\hphantom{(g :condition (o }:op2 (r2 / rain \\
\hphantom{(g :condition (o :op2 (r2 }:polarity -)))
\end{amr} \noindent
These clauses are often fronted. In which case, use the inverse \texttt{condition-of} (See also \texttt{cause} \see{cause}).
\begin{amr}{{If it rains, we will stay home}}
(r / rain \\
\hphantom{(r }:condition-of (s / stay \\
\hphantom{(r :condition-of (s }:theme (w / we) \\
\hphantom{(r :condition-of (s }:location (h / home)))
\end{amr} \noindent
%

%%%%%%%%%%%%%%%%%%%%%%%%%%%%%%%%%%%%%%%%%%%%%%%%%%%%%%%%%%%%%%%%%%%%%%%%%%%%%%%%%%%%%%%%%%%%%%%%%%%%%%
%%%%%%%%%%%%%%%%%%%%%%%%%%%%%%%%%%%%%%%%%%%%%%%%%%%%%%%%%%%%%%%%%%%%%%%%%%%%%%%%%%%%%%%%%%%%%%%%%%%%%%
\subsubsection{Concession}\label{concession}     % GREG
%%%%%%%%%%%%%%%%%%%%%%%%%%%%%%%%%%%%%%%%%%%%%%%%%%%%%%%%%%%%%%%%%%%%%%%%%%%%%%%%%%%%%%%%%%%%%%%%%%%%%%
%%%%%%%%%%%%%%%%%%%%%%%%%%%%%%%%%%%%%%%%%%%%%%%%%%%%%%%%%%%%%%%%%%%%%%%%%%%%%%%%%%%%%%%%%%%%%%%%%%%%%%

WISeR uses the role \texttt{concession} in the same way as AMR. It is used to represent concessive connectives such as \textit{although} and \textit{despite}.
\begin{amr}{The game continued although it rained\\The game continued despite the rain}
(c / continue \\
\hphantom{(c }:theme (g / game) \\
\hphantom{(c }:concession (r / rain))
\end{amr} \noindent
These clauses are often fronted, in which case you can use the inverse \texttt{concession-of}.
\begin{amr}{{Although it rained, the game continued\\Despite the rain, the game continued}}
(r / rain \\
\hphantom{(c }:concession-of (c / continue \\
\hphantom{(c :concession (r }:theme (g / game)))
\end{amr} \noindent
Sometimes \textit{but} is used concessively (see also \texttt{comparison} \see{comparison} for contrastive uses of \textit{but}).
\begin{amr}{Trade has developed rapidly but it still has potential}
(d / develop \\
\hphantom{(d }:theme (t / trade) \\
\hphantom{(d }:manner (r / rapid) \\
\hphantom{(d }:concession-of (h / have \\ 
\hphantom{(d :concession-of (h }:actor t \\ 
\hphantom{(d :concession-of (h }:theme (p / potential) \\
\hphantom{(d :concession-of (h }:mod (s / still)))
\end{amr} \noindent
%

%%%%%%%%%%%%%%%%%%%%%%%%%%%%%%%%%%%%%%%%%%%%%%%%%%%%%%%%%%%%%%%%%%%%%%%%%%%%%%%%%%%%%%%%%%%%%%%%%%%%%%
%%%%%%%%%%%%%%%%%%%%%%%%%%%%%%%%%%%%%%%%%%%%%%%%%%%%%%%%%%%%%%%%%%%%%%%%%%%%%%%%%%%%%%%%%%%%%%%%%%%%%%
\subsection{Mereology and Degrees}\label{mereology and degrees}
%%%%%%%%%%%%%%%%%%%%%%%%%%%%%%%%%%%%%%%%%%%%%%%%%%%%%%%%%%%%%%%%%%%%%%%%%%%%%%%%%%%%%%%%%%%%%%%%%%%%%%
%%%%%%%%%%%%%%%%%%%%%%%%%%%%%%%%%%%%%%%%%%%%%%%%%%%%%%%%%%%%%%%%%%%%%%%%%%%%%%%%%%%%%%%%%%%%%%%%%%%%%%

%%%%%%%%%%%%%%%%%%%%%%%%%%%%%%%%%%%%%%%%%%%%%%%%%%%%%%%%%%%%%%%%%%%%%%%%%%%%%%%%%%%%%%%%%%%%%%%%%%%%%%
%%%%%%%%%%%%%%%%%%%%%%%%%%%%%%%%%%%%%%%%%%%%%%%%%%%%%%%%%%%%%%%%%%%%%%%%%%%%%%%%%%%%%%%%%%%%%%%%%%%%%%
\subsubsection{Domain and Mod}\label{domain and mod}     % GREG
%%%%%%%%%%%%%%%%%%%%%%%%%%%%%%%%%%%%%%%%%%%%%%%%%%%%%%%%%%%%%%%%%%%%%%%%%%%%%%%%%%%%%%%%%%%%%%%%%%%%%%
%%%%%%%%%%%%%%%%%%%%%%%%%%%%%%%%%%%%%%%%%%%%%%%%%%%%%%%%%%%%%%%%%%%%%%%%%%%%%%%%%%%%%%%%%%%%%%%%%%%%%%

The roles \texttt{domain} and \texttt{mod} are inverses. The former is typically used in noun-copula-noun constructions.
\begin{amr}{They are birds}
(b / birds \\
\hphantom{(b }:domain (t / they))
\end{amr} \noindent
As well as in small clauses.
\begin{amr}{I consider him a friend}
(c / consider \\
\hphantom{(c }:actor (i / i) \\
\hphantom{(c }:theme (f / friend \\
\hphantom{(c :theme (s }:domain (h / he))
\end{amr} \noindent
\begin{amr}{They are considered traitors}
(c / consider \\
\hphantom{(c }:theme (p / person \\
\hphantom{(c :theme (p }:domain (t / they) \\
\hphantom{(c :theme (p }:actor-of (b / betray)))
\end{amr} \noindent
The role \texttt{mod} is typically used for nominal modifiers such as adjectives.
\begin{amr}{Vice president}
(p / president \\
\hphantom{(p }:mod (v / vice))
\end{amr} \noindent
As well as relative clauses in which the main predicate is a noun (i.e., when you need to use the inverse of \texttt{domain}).
\begin{amr}{The man who is a lawyer}
(m / man \\
\hphantom{(m }:mod (l / lawyer))
\end{amr} \noindent
It is important to note, however, that \texttt{mod} is not used for all adjectives. Since the concept \texttt{toy} could be a \texttt{theme} of the concept \texttt{new} (not \texttt{domain}) we use the inverse of \texttt{theme}, \texttt{theme-of}, not \texttt{mod}.
\begin{amr}{The new toy}
(t / toy \\
\hphantom{(t }:theme-of (n / new))
\end{amr} \noindent
Likewise for \texttt{weather} and \texttt{cold}.
\begin{amr}{The cold weather}
(w / weather \\
\hphantom{(w }:theme-of (c / cold))
\end{amr} \noindent
Consider also the following more complicated example.
\begin{amr}{My favorite dog\\The dog I favor}
(d / dog \\
\hphantom{(d }:theme-of (f / favor \\
\hphantom{(d :theme-of (f }:actor (i / i)))
\end{amr} \noindent
%

%%%%%%%%%%%%%%%%%%%%%%%%%%%%%%%%%%%%%%%%%%%%%%%%%%%%%%%%%%%%%%%%%%%%%%%%%%%%%%%%%%%%%%%%%%%%%%%%%%%%%%
%%%%%%%%%%%%%%%%%%%%%%%%%%%%%%%%%%%%%%%%%%%%%%%%%%%%%%%%%%%%%%%%%%%%%%%%%%%%%%%%%%%%%%%%%%%%%%%%%%%%%%
\subsubsection{Attribute}\label{attribute} % GREG
%%%%%%%%%%%%%%%%%%%%%%%%%%%%%%%%%%%%%%%%%%%%%%%%%%%%%%%%%%%%%%%%%%%%%%%%%%%%%%%%%%%%%%%%%%%%%%%%%%%%%%
%%%%%%%%%%%%%%%%%%%%%%%%%%%%%%%%%%%%%%%%%%%%%%%%%%%%%%%%%%%%%%%%%%%%%%%%%%%%%%%%%%%%%%%%%%%%%%%%%%%%%%

WISeR introduces the \texttt{attribute} role to account for a number of verb specific arguments, as well as providing a more intuitive description for some existing roles. The role \texttt{attribute} is used to annotate an argument which answers the question ``In what respect does an argument have, or change in, the property described?''.

Oftentimes, the \texttt{attribute} can appear redundant.
\begin{amr}{The man is short in stature}
(s / short \\
\hphantom{(s }:theme (m / man) \\
\hphantom{(s }:attribute (s2 / stature))
\end{amr} \noindent
\begin{amr}{The popcorn was free of charge}
(f / free \\
\hphantom{(f }:theme (p / popcorn) \\
\hphantom{(f }:attribute (c / charge))
\end{amr} \noindent
But this is not always the case.\footnote{For the sentence \textit{the man is rich in spirit}, PropBank would give \textit{the man} the thematic role goal, and \textit{spirit} the thematic role theme.}
\begin{amr}{The man grew in courage}
(g / grow \\
\hphantom{(g }:theme (m / man) \\
\hphantom{(g }:attribute (c / courage))
\end{amr} \noindent
\begin{amr}{The man is rich in spirit}
(r / rich \\
\hphantom{(f }:theme (m / man) \\
\hphantom{(f }:attribute (s / spirit))
\end{amr} \noindent
\begin{amr}{Silver's advance in price}
(a / advance \\
\hphantom{(a }:theme (s / silver) \\
\hphantom{(a }:attribute (p / price))
\end{amr} \noindent
Attributes are commonly introduced by the prepositions \textit{as} and \textit{in}, and they add more specific information about some feature of one of the arguments (typically the \texttt{theme}). This includes non-result state secondary predicates.
\begin{amr}{The woman was accredited as an expert}
(a / accredited \\
\hphantom{(a }:theme (w / woman) \\
\hphantom{(a }:attribute (e / expert))
\end{amr} \noindent
\begin{amr}{The girl was denounced as a fraud}
(d / denounce \\
\hphantom{(d }:theme (g / girl) \\
\hphantom{(d }:attribute (f / fraud))
\end{amr} \noindent
\begin{amr}{The girl employed the boy as a cleaner}
(e / employ \\
\hphantom{(e }:actor (g / girl) \\
\hphantom{(e }:theme (b / boy) \\
\hphantom{(e }:attribute (p / person \\
\hphantom{(e :attribute (p }:actor-of (c / clean)))
\end{amr} \noindent
\begin{amr}{Lying counts as a sin}
(c / count \\
\hphantom{(c }:theme (l / lie) \\
\hphantom{(c }:attribute (s / sin))
\end{amr} \noindent
%

%%%%%%%%%%%%%%%%%%%%%%%%%%%%%%%%%%%%%%%%%%%%%%%%%%%%%%%%%%%%%%%%%%%%%%%%%%%%%%%%%%%%%%%%%%%%%%%%%%%%%%
%%%%%%%%%%%%%%%%%%%%%%%%%%%%%%%%%%%%%%%%%%%%%%%%%%%%%%%%%%%%%%%%%%%%%%%%%%%%%%%%%%%%%%%%%%%%%%%%%%%%%%
\subsubsection{Quantity}\label{quantity}         % GREG
%%%%%%%%%%%%%%%%%%%%%%%%%%%%%%%%%%%%%%%%%%%%%%%%%%%%%%%%%%%%%%%%%%%%%%%%%%%%%%%%%%%%%%%%%%%%%%%%%%%%%%
%%%%%%%%%%%%%%%%%%%%%%%%%%%%%%%%%%%%%%%%%%%%%%%%%%%%%%%%%%%%%%%%%%%%%%%%%%%%%%%%%%%%%%%%%%%%%%%%%%%%%%

The relation \texttt{quantity} is used to annotate numerical amounts.
\begin{amr}{Three boys passed the exam}
(p / pass \\
\hphantom{(p }:actor (b / boy \\
\hphantom{(p :actor (b }:quant \textcolor{DarkBlue}{3}) \\
\hphantom{(p }:theme (e / exam))
\end{amr} \noindent
\begin{amr}{Several hundred apples}
(a / apples \\
\hphantom{(a }:quant (s / several \\
\hphantom{(a :quant (s }:op1 \textcolor{DarkBlue}{100}))
\end{amr} \noindent
\begin{amr}{Four out of five investors lost money}
(l / lost \\
\hphantom{(l }:actor (p / person \\
\hphantom{(l :actor (p }:actor-of (i / invest) \\
\hphantom{(l :actor (p }:quant \textcolor{DarkBlue}{4} \\
\hphantom{(l :actor (p }:subset-of (p2 / person \\
\hphantom{(l :actor (p :subset-of (p2 }:actor-of (i2 / invest) \\
\hphantom{(l :actor (p :subset-of (p2 }:quant \textcolor{DarkBlue}{5})) \\
\hphantom{(l }:theme (m / money))
\end{amr} \noindent
It is also used to specify distance quantities and temporal quantities (See \texttt{extent} \see{extent}, \texttt{duration} \see{duration} and \texttt{range} \see{range}).

%%%%%%%%%%%%%%%%%%%%%%%%%%%%%%%%%%%%%%%%%%%%%%%%%%%%%%%%%%%%%%%%%%%%%%%%%%%%%%%%%%%%%%%%%%%%%%%%%%%%%%
%%%%%%%%%%%%%%%%%%%%%%%%%%%%%%%%%%%%%%%%%%%%%%%%%%%%%%%%%%%%%%%%%%%%%%%%%%%%%%%%%%%%%%%%%%%%%%%%%%%%%%
\subsubsection{Degree}\label{degree}             % GREG
%%%%%%%%%%%%%%%%%%%%%%%%%%%%%%%%%%%%%%%%%%%%%%%%%%%%%%%%%%%%%%%%%%%%%%%%%%%%%%%%%%%%%%%%%%%%%%%%%%%%%%
%%%%%%%%%%%%%%%%%%%%%%%%%%%%%%%%%%%%%%%%%%%%%%%%%%%%%%%%%%%%%%%%%%%%%%%%%%%%%%%%%%%%%%%%%%%%%%%%%%%%%%

The \texttt{degree} role is used to introduce intensifiers like \textit{very}, and \textit{extremely} as well as ``downtoners'' like \textit{somewhat} and \textit{relatively}.
\begin{amr}{The girl is very tall}
(t / tall \\
\hphantom{(t }:theme (g / girl) \\
\hphantom{(t }:degree (v / very))
\end{amr} \noindent
\begin{amr}{The girl is too tall}
(t / tall \\
\hphantom{(t }:theme (g / girl) \\
\hphantom{(t }:degree (t / too))
\end{amr} \noindent
It is also used in comparatives and superlatives (See also \texttt{comparison} \see{comparison}).
\begin{amr}{The girl is the best}
(g / good \\
\hphantom{(g }:theme (g2 / girl) \\
\hphantom{(g }:degree (m / most))
\end{amr} \noindent
%

%%%%%%%%%%%%%%%%%%%%%%%%%%%%%%%%%%%%%%%%%%%%%%%%%%%%%%%%%%%%%%%%%%%%%%%%%%%%%%%%%%%%%%%%%%%%%%%%%%%%%%
%%%%%%%%%%%%%%%%%%%%%%%%%%%%%%%%%%%%%%%%%%%%%%%%%%%%%%%%%%%%%%%%%%%%%%%%%%%%%%%%%%%%%%%%%%%%%%%%%%%%%%
\subsubsection{Extent}\label{extent} % GREG
%%%%%%%%%%%%%%%%%%%%%%%%%%%%%%%%%%%%%%%%%%%%%%%%%%%%%%%%%%%%%%%%%%%%%%%%%%%%%%%%%%%%%%%%%%%%%%%%%%%%%%
%%%%%%%%%%%%%%%%%%%%%%%%%%%%%%%%%%%%%%%%%%%%%%%%%%%%%%%%%%%%%%%%%%%%%%%%%%%%%%%%%%%%%%%%%%%%%%%%%%%%%%

The role \texttt{extent} is not to be confused with \texttt{degree}. This role is often used to quantify a predicate.
\begin{amr}{The road goes on forever}
(g / go-on \\
\hphantom{(g }:theme (r / road \\
\hphantom{(g }:extent (f / forever))
\end{amr} \noindent
\begin{amr}{The boy grew 3 inches}
(g / grow \\
\hphantom{(g }:theme (b / boy \\
\hphantom{(g }:extent (d / distance-quantity \\
\hphantom{(g :extent (d }:unit (i / inches \\
\hphantom{(g :extent (d :unit (i }:quant \textcolor{DarkBlue}{3})))
\end{amr} \noindent
We will also use this relation to introduce a measure phrase in comparative constructions (See also \texttt{comparison} \see{comparison}).

%%%%%%%%%%%%%%%%%%%%%%%%%%%%%%%%%%%%%%%%%%%%%%%%%%%%%%%%%%%%%%%%%%%%%%%%%%%%%%%%%%%%%%%%%%%%%%%%%%%%%%
%%%%%%%%%%%%%%%%%%%%%%%%%%%%%%%%%%%%%%%%%%%%%%%%%%%%%%%%%%%%%%%%%%%%%%%%%%%%%%%%%%%%%%%%%%%%%%%%%%%%%%
\subsubsection{Comparison}\label{comparison}            % GREG
%%%%%%%%%%%%%%%%%%%%%%%%%%%%%%%%%%%%%%%%%%%%%%%%%%%%%%%%%%%%%%%%%%%%%%%%%%%%%%%%%%%%%%%%%%%%%%%%%%%%%%
%%%%%%%%%%%%%%%%%%%%%%%%%%%%%%%%%%%%%%%%%%%%%%%%%%%%%%%%%%%%%%%%%%%%%%%%%%%%%%%%%%%%%%%%%%%%%%%%%%%%%%

The annotations in the AMR 3.0 corpus follow the suggestions in \cite{bonial2018abstract}. As such, we adopt these suggestions for now, modulo the discarding of numbered \texttt{ARG}s.\footnote{WISeR aims to seek potential improvements on this work in the future.} Comparatives are represented using a reification of the \texttt{degree} relation, \texttt{have-degree}. Since WISeR does not use numbered \texttt{ARG}s, we introduce a \texttt{comparison} relation. The \texttt{comparison} role is given to arguments which something is being compared to or contrasted with.
\begin{amr}{The girl is taller than the boy\\The girl is taller than the boy is}
(h / have-degree \\
\hphantom{(t }:theme (g / girl) \\
\hphantom{(t }:attribute (t / tall) \\
\hphantom{(t }:degree (m / more) \\
\hphantom{(t }:comparison (b / boy))
\end{amr} \noindent
A full list of the relations used are as follows.\footnote{Notice that we collapse three of \citeauthor{bonial2018abstract}'s numbered \texttt{ARG} roles into one \texttt{comparison} role. We do this for several reasons: (i) they are all responsible for introducing a point of comparison, (ii) they never co-occur, (iii) the choice of numbered \texttt{ARG} depends entirely on the value of the \texttt{degree} role. As such, we may be able to get away with assuming that interpretation of \texttt{comparison} simply depends on the value of the \texttt{degree} (e.g., \texttt{more}, \texttt{most}, \texttt{too}, etc.).}

\begin{table}[!h]
    \centering
    \begin{tabular}{| c || c |}
        \hline
        \texttt{theme} & entity characterized by attribute \\ \hline
        \texttt{attribute} & attribute (e.g. tall) \\ \hline
        \texttt{degree} & degree itself (e.g. more/most, less/least, equal) \\  \hline
        \texttt{comparison} & compared-to  \\  \hline
        \texttt{comparison} &  reference to superset \\  \hline
        \texttt{comparison} & consequence, result of degree \\  \hline
    \end{tabular}
    \caption{List of arguments for \texttt{have-degree}}
    \label{tab:degree-args}
\end{table}

\noindent
Below is an example of a superlative with a \texttt{comparison} argument.
\begin{amr}{The girl is the tallest of her friends}
(h / have-degree \\
\hphantom{(t }:theme (g / girl) \\
\hphantom{(t }:attribute (t / tall) \\
\hphantom{(t }:degree (m / most) \\
\hphantom{(t }:comparison (f / friend \\
\hphantom{(t :comparison (f }:poss g))
\end{amr} \noindent
The following is an example of a `degree consequence' construction.
\begin{amr}{The girl is too tall to sit comfortably}
(h / have-degree \\
\hphantom{(t }:theme (g / girl) \\
\hphantom{(t }:attribute (t / tall) \\
\hphantom{(t }:degree (t2 / too) \\
\hphantom{(t }:comparison (s / sit \\
\hphantom{(t :comparison (s }:theme g \\
\hphantom{(t :comparison (s }:manner (c / comfort)))
\end{amr} \noindent
Notice that the above sentence would typically be said when the girl is unable to sit comfortably (i.e., the consequence clause is non-veridical). However, rather than inserting negation or a modal concept here, \cite{bonial2018abstract} leave this representation as it is.\footnote{They note that sentences such as \textit{the man was too drunk to drive} do not always entail that the man didn't drive.} In later versions, WISeR aims to make improvements in this respect.

Finally, we use \texttt{comparison} for certain verbal arguments, such as the second prototypical patient/theme argument assigned by PropBank to the verb \textit{correlate}.
\begin{amr}{Life expectancy correlates with wages}
(c / correlate \\
\hphantom{(c }:theme (e / expect \\
\hphantom{(c :theme (e }:theme (l / live)) \\
\hphantom{(c }:comparison (w / wage))
\end{amr} \noindent
Likewise, PropBank assigns the first argument of \textit{similar} an agent role and the second a patient/theme role. However, neither argument can reasonably be called an agent. The addition of \texttt{comparison} allows us to rectify this.
\begin{amr}{The girl is similar to the boy in height}
(s / similar \\
\hphantom{(s }:theme (g / girl) \\
\hphantom{(s }:comparison (b / boy) \\
\hphantom{(s }:attribute (h / height))
\end{amr} \noindent
We also use comparison for arguments of \texttt{contrast} and contrastive connectives such as \textit{but} (following the annotation of contrastive \textit{but} in AMR 3.0).
\begin{amr}{{The boy likes it, but the girl does not.}}
(l / like \\
\hphantom{(l }:actor (b / boy) \\
\hphantom{(l }:theme (i / it) \\
\hphantom{(l }:theme-of (c / contrast \\
\hphantom{(l :theme-of (c }:comparison (l2 / like \\
\hphantom{(l :theme-of (c :comparison (l2 }:actor (g / girl) \\
\hphantom{(l :theme-of (c :comparison (l2 }:theme i \\
\hphantom{(l :theme-of (c :comparison (l2 }:polarity -))) \\
\end{amr} \noindent
%

%%%%%%%%%%%%%%%%%%%%%%%%%%%%%%%%%%%%%%%%%%%%%%%%%%%%%%%%%%%%%%%%%%%%%%%%%%%%%%%%%%%%%%%%%%%%%%%%%%%%%%
%%%%%%%%%%%%%%%%%%%%%%%%%%%%%%%%%%%%%%%%%%%%%%%%%%%%%%%%%%%%%%%%%%%%%%%%%%%%%%%%%%%%%%%%%%%%%%%%%%%%%%
\subsubsection{Possession}\label{possession}       %    Lydia 
%%%%%%%%%%%%%%%%%%%%%%%%%%%%%%%%%%%%%%%%%%%%%%%%%%%%%%%%%%%%%%%%%%%%%%%%%%%%%%%%%%%%%%%%%%%%%%%%%%%%%%
%%%%%%%%%%%%%%%%%%%%%%%%%%%%%%%%%%%%%%%%%%%%%%%%%%%%%%%%%%%%%%%%%%%%%%%%%%%%%%%%%%%%%%%%%%%%%%%%%%%%%%

The \texttt{poss} relation is used to represent ownership or possession.
\begin{amr}{{He loved his children}}
(l / love \\
\hphantom{(l }:actor (h / he) \\
\hphantom{(l }:benefactive (c / children \\
\hphantom{(l :benefactive (c }:poss h))
\end{amr} \noindent
Note that \texttt{poss} is different from \texttt{part-of}, as it shows ownership not the relationship between two parts of one thing.
\begin{amr}{{The sailor's boat}}
(b / boat \\
\hphantom{(b }:poss (s / sailor))
\end{amr} \noindent
\begin{amr}{{The boat's sail}}
(s / sail \\
\hphantom{(s }:part-of (b / boat))
\end{amr} \noindent
%

%%%%%%%%%%%%%%%%%%%%%%%%%%%%%%%%%%%%%%%%%%%%%%%%%%%%%%%%%%%%%%%%%%%%%%%%%%%%%%%%%%%%%%%%%%%%%%%%%%%%%%
%%%%%%%%%%%%%%%%%%%%%%%%%%%%%%%%%%%%%%%%%%%%%%%%%%%%%%%%%%%%%%%%%%%%%%%%%%%%%%%%%%%%%%%%%%%%%%%%%%%%%%
\subsubsection{Part-of and Consist-of}\label{part-of and consist-of}      %    Greg
%%%%%%%%%%%%%%%%%%%%%%%%%%%%%%%%%%%%%%%%%%%%%%%%%%%%%%%%%%%%%%%%%%%%%%%%%%%%%%%%%%%%%%%%%%%%%%%%%%%%%%
%%%%%%%%%%%%%%%%%%%%%%%%%%%%%%%%%%%%%%%%%%%%%%%%%%%%%%%%%%%%%%%%%%%%%%%%%%%%%%%%%%%%%%%%%%%%%%%%%%%%%%

We use \texttt{consist-of} to represent the substance which an instance of a concept is composed from.
\begin{amr}{The gold watch}
(w / watch \\
\hphantom{(w }:consist-of (g / gold))
\end{amr} \noindent
We can also use it to cover some verb specific roles such as that of \texttt{compose}.\footnote{This receives the  vn-role `material' in the PropBank frame.}
\begin{amr}{The team is composed of players}
(c / compose \\
\hphantom{(c }:theme (t / team) \\
\hphantom{(c }:consist-of (p / player))
\end{amr} \noindent
This can be read `the composition of the team consists of players'.

The \texttt{part-of} relation is used when representing a part of an entity.
\begin{amr}{The engine of the car\\The car's engine}
(e / engine \\
\hphantom{(e }:part-of (c / car))
\end{amr} \noindent
\begin{amr}{The boy's leg}
(l / leg \\
\hphantom{(c }:part-of (b / boy))
\end{amr} \noindent
\begin{amr}{The south of France}
(s / south \\
\hphantom{(s }:part-of (c / country :wiki "France" \\
\hphantom{(s :part-of (c }:name (n / name :op1 "France")))
\end{amr} \noindent
%

%%%%%%%%%%%%%%%%%%%%%%%%%%%%%%%%%%%%%%%%%%%%%%%%%%%%%%%%%%%%%%%%%%%%%%%%%%%%%%%%%%%%%%%%%%%%%%%%%%%%%%
%%%%%%%%%%%%%%%%%%%%%%%%%%%%%%%%%%%%%%%%%%%%%%%%%%%%%%%%%%%%%%%%%%%%%%%%%%%%%%%%%%%%%%%%%%%%%%%%%%%%%%
\subsubsection{Subevent}\label{subevent}         %  Lydia 
%%%%%%%%%%%%%%%%%%%%%%%%%%%%%%%%%%%%%%%%%%%%%%%%%%%%%%%%%%%%%%%%%%%%%%%%%%%%%%%%%%%%%%%%%%%%%%%%%%%%%%
%%%%%%%%%%%%%%%%%%%%%%%%%%%%%%%%%%%%%%%%%%%%%%%%%%%%%%%%%%%%%%%%%%%%%%%%%%%%%%%%%%%%%%%%%%%%%%%%%%%%%%

The \texttt{subevent} relation is used to describe the larger event of which the event in question is a part. It is often introduced with the phrase \textit{in which}.
\begin{amr}{A massive bombardment in which 300 missiles rained on the capital}
(b / bombard \\
\hphantom{(b }:mod (m / massive) \\
\hphantom{(b }:subevent (r / rain \\
\hphantom{(b :subevent (r }:theme (m / missiles \\
\hphantom{(b :subevent (r :theme (m }:quant \textcolor{DarkBlue}{300}) \\
\hphantom{(b :subevent (r }:location (c / capital) \\
\hphantom{(b :subevent (r }:direction (d / down))))
\end{amr} \noindent
It contextualizes the event as part of an overarching event.
\begin{amr}{The speakers left on the final day of the conference}
(l / leave \\
\hphantom{(l }:actor (s / speakers) \\
\hphantom{(l }:time (d / day \\
\hphantom{(l :time (d }:mod (f / final) \\
\hphantom{(l :time (d }:subevent (c / conference)))
\end{amr} \noindent
%

%%%%%%%%%%%%%%%%%%%%%%%%%%%%%%%%%%%%%%%%%%%%%%%%%%%%%%%%%%%%%%%%%%%%%%%%%%%%%%%%%%%%%%%%%%%%%%%%%%%%%%
%%%%%%%%%%%%%%%%%%%%%%%%%%%%%%%%%%%%%%%%%%%%%%%%%%%%%%%%%%%%%%%%%%%%%%%%%%%%%%%%%%%%%%%%%%%%%%%%%%%%%%
\subsection{Operators}\label{operators}
%%%%%%%%%%%%%%%%%%%%%%%%%%%%%%%%%%%%%%%%%%%%%%%%%%%%%%%%%%%%%%%%%%%%%%%%%%%%%%%%%%%%%%%%%%%%%%%%%%%%%%
%%%%%%%%%%%%%%%%%%%%%%%%%%%%%%%%%%%%%%%%%%%%%%%%%%%%%%%%%%%%%%%%%%%%%%%%%%%%%%%%%%%%%%%%%%%%%%%%%%%%%%

The WISeR roles described in this section are adopted wholesale from the AMR guidelines. Annotators with experience converting text into AMR can safely skip this section.

%%%%%%%%%%%%%%%%%%%%%%%%%%%%%%%%%%%%%%%%%%%%%%%%%%%%%%%%%%%%%%%%%%%%%%%%%%%%%%%%%%%%%%%%%%%%%%%%%%%%%%
%%%%%%%%%%%%%%%%%%%%%%%%%%%%%%%%%%%%%%%%%%%%%%%%%%%%%%%%%%%%%%%%%%%%%%%%%%%%%%%%%%%%%%%%%%%%%%%%%%%%%%
\subsubsection{Op}\label{op}       %    Greg 
%%%%%%%%%%%%%%%%%%%%%%%%%%%%%%%%%%%%%%%%%%%%%%%%%%%%%%%%%%%%%%%%%%%%%%%%%%%%%%%%%%%%%%%%%%%%%%%%%%%%%%
%%%%%%%%%%%%%%%%%%%%%%%%%%%%%%%%%%%%%%%%%%%%%%%%%%%%%%%%%%%%%%%%%%%%%%%%%%%%%%%%%%%%%%%%%%%%%%%%%%%%%%

As in AMR, \texttt{opx} roles are used in conjunctions and disjunctions.
\begin{amr}{The boy and the girl swam}
(s / swim \\
\hphantom{(s }:actor (a / and \\
\hphantom{(s :actor (a }:op1 (b / boy) \\
\hphantom{(s :actor (a }:op2 (g / girl))
\end{amr} \noindent
As well as in spatial and temporal arguments.
\begin{amr}{The boy sang 10 minutes ago}
(s / sing \\
\hphantom{(s }:actor (b / boy) \\
\hphantom{(s }:time (b2 / before \\
\hphantom{(s :time (b2 }:op1 (n / now) \\
\hphantom{(s :time (b2 }:quant (t / temporal-quantity \\
\hphantom{(s :time (b2 :quant (t }:unit (m / minutes) \\
\hphantom{(s :time (b2 :quant (t }:quant \textcolor{DarkBlue}{10})))
\end{amr} \noindent
And for named entities.
\begin{amr}{The Titanic}
(s / ship \\
\hphantom{(s }:wiki "RMS\_Titanic" \\
\hphantom{(s }:name (n / name) \\
\hphantom{(s :name (n }:op1 "Titanic"))
\end{amr} \noindent
For more uses of \texttt{opx}, refer to the AMR guidelines.

%%%%%%%%%%%%%%%%%%%%%%%%%%%%%%%%%%%%%%%%%%%%%%%%%%%%%%%%%%%%%%%%%%%%%%%%%%%%%%%%%%%%%%%%%%%%%%%%%%%%%%
%%%%%%%%%%%%%%%%%%%%%%%%%%%%%%%%%%%%%%%%%%%%%%%%%%%%%%%%%%%%%%%%%%%%%%%%%%%%%%%%%%%%%%%%%%%%%%%%%%%%%%
\subsubsection{Polarity}\label{polarity}      %    Lydia 
%%%%%%%%%%%%%%%%%%%%%%%%%%%%%%%%%%%%%%%%%%%%%%%%%%%%%%%%%%%%%%%%%%%%%%%%%%%%%%%%%%%%%%%%%%%%%%%%%%%%%%
%%%%%%%%%%%%%%%%%%%%%%%%%%%%%%%%%%%%%%%%%%%%%%%%%%%%%%%%%%%%%%%%%%%%%%%%%%%%%%%%%%%%%%%%%%%%%%%%%%%%%%
The \texttt{polarity} relation is used to evaluate the logical truth value of the statement and can be used to negate sentences. This relation is a binary value.
\begin{amr}{The boy doesn't go}
(g / go \\
\hphantom{(g }:actor (b / boy) \\
\hphantom{(g }:polarity -)
\end{amr} \noindent
This role negates the predicate under which it is immediately nested. Consider the following example in contrast to the first.
\begin{amr}{It is not the boy who goes}
(g / go \\
\hphantom{(g }:actor (b / boy) \\
\hphantom{(g :actor (b }:polarity -)
\end{amr} \noindent
%

%%%%%%%%%%%%%%%%%%%%%%%%%%%%%%%%%%%%%%%%%%%%%%%%%%%%%%%%%%%%%%%%%%%%%%%%%%%%%%%%%%%%%%%%%%%%%%%%%%%%%%
%%%%%%%%%%%%%%%%%%%%%%%%%%%%%%%%%%%%%%%%%%%%%%%%%%%%%%%%%%%%%%%%%%%%%%%%%%%%%%%%%%%%%%%%%%%%%%%%%%%%%%
\subsubsection{Polite}\label{polite}         %    Lydia 
%%%%%%%%%%%%%%%%%%%%%%%%%%%%%%%%%%%%%%%%%%%%%%%%%%%%%%%%%%%%%%%%%%%%%%%%%%%%%%%%%%%%%%%%%%%%%%%%%%%%%%
%%%%%%%%%%%%%%%%%%%%%%%%%%%%%%%%%%%%%%%%%%%%%%%%%%%%%%%%%%%%%%%%%%%%%%%%%%%%%%%%%%%%%%%%%%%%%%%%%%%%%%

The \texttt{polite} role is used to annotate politeness markers. This role has a binary value.
\begin{amr}{We'd ask you to please leave}
(a / ask \\
\hphantom{(a }:actor (w / we) \\
\hphantom{(a }:benefactive (y / you) \\
\hphantom{(a }:theme (l / leave \\
\hphantom{(a :theme (l }:actor y \\
\hphantom{(a :theme (l }:polite +))
\end{amr} \noindent
%
% Are there any examples of polite - ?

%%%%%%%%%%%%%%%%%%%%%%%%%%%%%%%%%%%%%%%%%%%%%%%%%%%%%%%%%%%%%%%%%%%%%%%%%%%%%%%%%%%%%%%%%%%%%%%%%%%%%%
%%%%%%%%%%%%%%%%%%%%%%%%%%%%%%%%%%%%%%%%%%%%%%%%%%%%%%%%%%%%%%%%%%%%%%%%%%%%%%%%%%%%%%%%%%%%%%%%%%%%%%
\subsubsection{Mode}\label{mode}           %    Lydia 
%%%%%%%%%%%%%%%%%%%%%%%%%%%%%%%%%%%%%%%%%%%%%%%%%%%%%%%%%%%%%%%%%%%%%%%%%%%%%%%%%%%%%%%%%%%%%%%%%%%%%%
%%%%%%%%%%%%%%%%%%%%%%%%%%%%%%%%%%%%%%%%%%%%%%%%%%%%%%%%%%%%%%%%%%%%%%%%%%%%%%%%%%%%%%%%%%%%%%%%%%%%%%

The \texttt{mode} role describes the mood of the sentence and the intentions of the speakers. It can mark an imperative.
\begin{amr}{Let's go!}
(g / go \\
\hphantom{(g }:actor (w / we) \\
\hphantom{(g }:mode imperative)
\end{amr} \noindent
\begin{amr}{Wait here!}
(w / wait \\
\hphantom{(g }:actor (y / you) \\
\hphantom{(g }:location (h / here) \\
\hphantom{(g }:mode imperative)
\end{amr} \noindent
Or an expressive.
\begin{amr}{Wow!}
(w / wow \\
\hphantom{(w }:mode expressive)
\end{amr} \noindent
%

%%%%%%%%%%%%%%%%%%%%%%%%%%%%%%%%%%%%%%%%%%%%%%%%%%%%%%%%%%%%%%%%%%%%%%%%%%%%%%%%%%%%%%%%%%%%%%%%%%%%%%
%%%%%%%%%%%%%%%%%%%%%%%%%%%%%%%%%%%%%%%%%%%%%%%%%%%%%%%%%%%%%%%%%%%%%%%%%%%%%%%%%%%%%%%%%%%%%%%%%%%%%%
\subsubsection{Example}\label{example}             %    Lydia 
%%%%%%%%%%%%%%%%%%%%%%%%%%%%%%%%%%%%%%%%%%%%%%%%%%%%%%%%%%%%%%%%%%%%%%%%%%%%%%%%%%%%%%%%%%%%%%%%%%%%%%
%%%%%%%%%%%%%%%%%%%%%%%%%%%%%%%%%%%%%%%%%%%%%%%%%%%%%%%%%%%%%%%%%%%%%%%%%%%%%%%%%%%%%%%%%%%%%%%%%%%%%%

The \texttt{example} role introduces something which is an example of a concept
\begin{amr}{The family vacations in resort spots like the beach}
(v / vacation \\
\hphantom{(v }:actor (f / family) \\
\hphantom{(v }:location (s / spots  \\
\hphantom{(v :location (s }:mod (r / resort) \\
\hphantom{(v :location (s }:example (b / beach))) \\
\end{amr} \noindent
\begin{amr}{I like music such as country and rock}
(l / like \\
\hphantom{(l }:actor (i / i) \\
\hphantom{(l }:theme (m / music  \\
\hphantom{(l :theme (m }:example (a / and \\
\hphantom{(l :theme (m :example (a }:op1 (c / country) \\
\hphantom{(l :theme (m :example (a }:op1 (r / rock))))
\end{amr} \noindent
%

%%%%%%%%%%%%%%%%%%%%%%%%%%%%%%%%%%%%%%%%%%%%%%%%%%%%%%%%%%%%%%%%%%%%%%%%%%%%%%%%%%%%%%%%%%%%%%%%%%%%%%
%%%%%%%%%%%%%%%%%%%%%%%%%%%%%%%%%%%%%%%%%%%%%%%%%%%%%%%%%%%%%%%%%%%%%%%%%%%%%%%%%%%%%%%%%%%%%%%%%%%%%%
\subsubsection{Name}\label{name}        %    Lydia 
%%%%%%%%%%%%%%%%%%%%%%%%%%%%%%%%%%%%%%%%%%%%%%%%%%%%%%%%%%%%%%%%%%%%%%%%%%%%%%%%%%%%%%%%%%%%%%%%%%%%%%
%%%%%%%%%%%%%%%%%%%%%%%%%%%%%%%%%%%%%%%%%%%%%%%%%%%%%%%%%%%%%%%%%%%%%%%%%%%%%%%%%%%%%%%%%%%%%%%%%%%%%%

The \texttt{name} role provides a concept's name.
\begin{amr}{The family's dog Snoopy barked}
(b / bark \\
\hphantom{(b }:actor (d / dog \\
\hphantom{(b :actor (d }:poss (f / family) \\
\hphantom{(b :actor (d }:name (n / name \\
\hphantom{(b :actor (d :name (n }:op1 "Snoopy")))
\end{amr} \noindent
%

%%%%%%%%%%%%%%%%%%%%%%%%%%%%%%%%%%%%%%%%%%%%%%%%%%%%%%%%%%%%%%%%%%%%%%%%%%%%%%%%%%%%%%%%%%%%%%%%%%%%%%
%%%%%%%%%%%%%%%%%%%%%%%%%%%%%%%%%%%%%%%%%%%%%%%%%%%%%%%%%%%%%%%%%%%%%%%%%%%%%%%%%%%%%%%%%%%%%%%%%%%%%%
\subsubsection{Age}\label{age}           %    Lydia 
%%%%%%%%%%%%%%%%%%%%%%%%%%%%%%%%%%%%%%%%%%%%%%%%%%%%%%%%%%%%%%%%%%%%%%%%%%%%%%%%%%%%%%%%%%%%%%%%%%%%%%
%%%%%%%%%%%%%%%%%%%%%%%%%%%%%%%%%%%%%%%%%%%%%%%%%%%%%%%%%%%%%%%%%%%%%%%%%%%%%%%%%%%%%%%%%%%%%%%%%%%%%%

The \texttt{age} role provides an entity's age.
\begin{amr}{The 38 year old man injured his leg}
(i / injure \\
\hphantom{(i }:actor (m / man \\
\hphantom{(i :actor (m }:age (t / temporal-quantity \\
\hphantom{(i :actor (m :age (t }:quant 38 \\
\hphantom{(i :actor (m :age (t }:unit (y / year))) \\
\hphantom{(i }:theme (l / leg \\
\hphantom{(i :theme (l }:part-of m)) 
\end{amr} \noindent
%

%%%%%%%%%%%%%%%%%%%%%%%%%%%%%%%%%%%%%%%%%%%%%%%%%%%%%%%%%%%%%%%%%%%%%%%%%%%%%%%%%%%%%%%%%%%%%%%%%%%%%%
%%%%%%%%%%%%%%%%%%%%%%%%%%%%%%%%%%%%%%%%%%%%%%%%%%%%%%%%%%%%%%%%%%%%%%%%%%%%%%%%%%%%%%%%%%%%%%%%%%%%%%
\subsubsection{Value and Ord}\label{ord and value}              % GREG
%%%%%%%%%%%%%%%%%%%%%%%%%%%%%%%%%%%%%%%%%%%%%%%%%%%%%%%%%%%%%%%%%%%%%%%%%%%%%%%%%%%%%%%%%%%%%%%%%%%%%%
%%%%%%%%%%%%%%%%%%%%%%%%%%%%%%%%%%%%%%%%%%%%%%%%%%%%%%%%%%%%%%%%%%%%%%%%%%%%%%%%%%%%%%%%%%%%%%%%%%%%%%

The role \texttt{value} is used for specifying the numerical value of an entity.
\begin{amr}{Ninety-nine percent\\99\%}
(p / percentage-entity \\
\hphantom{(p }:value \textcolor{DarkBlue}{99})
\end{amr} \noindent
While \texttt{ord} is used for ordinal numbers (i.e., 1st, 2nd, 3rd, etc.)
\begin{amr}{The second planet}
(p / planet \\
\hphantom{(p }:ord (o / ordinal-entity \\
\hphantom{(p :ord (o }:value \textcolor{DarkBlue}{2}))
\end{amr} \noindent
%

%%%%%%%%%%%%%%%%%%%%%%%%%%%%%%%%%%%%%%%%%%%%%%%%%%%%%%%%%%%%%%%%%%%%%%%%%%%%%%%%%%%%%%%%%%%%%%%%%%%%%%
%%%%%%%%%%%%%%%%%%%%%%%%%%%%%%%%%%%%%%%%%%%%%%%%%%%%%%%%%%%%%%%%%%%%%%%%%%%%%%%%%%%%%%%%%%%%%%%%%%%%%%
\subsubsection{Unit}\label{unit}                 % GREG
%%%%%%%%%%%%%%%%%%%%%%%%%%%%%%%%%%%%%%%%%%%%%%%%%%%%%%%%%%%%%%%%%%%%%%%%%%%%%%%%%%%%%%%%%%%%%%%%%%%%%%
%%%%%%%%%%%%%%%%%%%%%%%%%%%%%%%%%%%%%%%%%%%%%%%%%%%%%%%%%%%%%%%%%%%%%%%%%%%%%%%%%%%%%%%%%%%%%%%%%%%%%%

The \texttt{unit} relation is used, often with \texttt{quantity}, to denote the measurement of a quantity. 
\begin{amr}{She had planned her wedding for ten years}
(p / plan \\
\hphantom{(b }:actor (s / she) \\
\hphantom{(b }:theme (w / wedding) \\
\hphantom{(b }:duration (t / temporal-quantity \\
\hphantom{(b :duration (t }:quantity \textcolor{DarkBlue}{10} \\
\hphantom{(b :duration (t }:unit (y / year)))
\end{amr} \noindent
Units also don't have to be scientifically measured units.
\begin{amr}{a dozen bottles of water}
(w / water \\
\hphantom{(w }:quantity (d / dozen) \\
\hphantom{(w }:unit (b / bottle))
\end{amr} \noindent
We also must be explicit about what we are measuring when we use units. In the below example, without the \texttt{weight-quantity} predicate the meaning representation would be under specified.
\begin{amr}{The couple bought 4 pounds of rice}
(b / buy \\
\hphantom{(b }:actor (c / couple) \\
\hphantom{(b }:theme (r / rice \\
\hphantom{(b :theme (r }:quantity (w / weight-quantity \\
\hphantom{(b :theme (r :quantity (w }:quantity \textcolor{DarkBlue}{4} \\
\hphantom{(b :theme (r :quantity (w }:unit (p / pound)))
\end{amr} \noindent
Similarly, we use \texttt{x-quantity} for other measurements such as volume for mass nouns.
\begin{amr}{The couple bought 2 gallons of milk}
(b / buy \\
\hphantom{(b }:actor (c / couple) \\
\hphantom{(b }:theme (m / milk \\
\hphantom{(b :theme (r }:quantity (v / volume-quantity \\
\hphantom{(b :theme (r :quantity (w }:quantity \textcolor{DarkBlue}{2} \\
\hphantom{(b :theme (r :quantity (w }:unit (g / gallon)))
\end{amr} \noindent

%%%%%%%%%%%%%%%%%%%%%%%%%%%%%%%%%%%%%%%%%%%%%%%%%%%%%%%%%%%%%%%%%%%%%%%%%%%%%%%%%%%%%%%%%%%%%%%%%%%%%%
%%%%%%%%%%%%%%%%%%%%%%%%%%%%%%%%%%%%%%%%%%%%%%%%%%%%%%%%%%%%%%%%%%%%%%%%%%%%%%%%%%%%%%%%%%%%%%%%%%%%%%
\subsubsection{List}\label{list}                  % Lydia 
%%%%%%%%%%%%%%%%%%%%%%%%%%%%%%%%%%%%%%%%%%%%%%%%%%%%%%%%%%%%%%%%%%%%%%%%%%%%%%%%%%%%%%%%%%%%%%%%%%%%%%
%%%%%%%%%%%%%%%%%%%%%%%%%%%%%%%%%%%%%%%%%%%%%%%%%%%%%%%%%%%%%%%%%%%%%%%%%%%%%%%%%%%%%%%%%%%%%%%%%%%%%%
The \texttt{list} relation is used to enumerate a list of items.
%example??
%
\begin{amr}{{She believed she lived in the best city- one, everyone was friendly; two, the weather was perfect; and three, the food was delicious}}
(m / multi-sentence \\
\hphantom{(e }:snt1 (b / believe \\
\hphantom{(e :snt1 (b }:actor (s / she) \\
\hphantom{(e :snt1 (b }:theme (r / reside \\
\hphantom{(e :snt1 (b :theme (l }:theme s \\
\hphantom{(e :snt1 (b :theme (l }:location (c / city \\
\hphantom{(e :snt1 (b :theme (l :location (c }:mod (b / best)))) \\
\hphantom{(e }:snt2 (f / friendly \\
\hphantom{(e :snt2 (f }:list 1 \\
\hphantom{(e :snt2 (f }:actor-of (e / everyone)) \\
\hphantom{(e }:snt3 (p / perfect \\
\hphantom{(e :snt3 (f }:list 2 \\
\hphantom{(e :snt3 (f }:theme-of (w / weather)) \\
\hphantom{(e }:snt4 (d / delicious \\
\hphantom{(e :snt4 (f }:list 3 \\
\hphantom{(e :snt4 (f }:theme-of (f / food)))
\end{amr} \noindent
%

%%%%%%%%%%%%%%%%%%%%%%%%%%%%%%%%%%%%%%%%%%%%%%%%%%%%%%%%%%%%%%%%%%%%%%%%%%%%%%%%%%%%%%%%%%%%%%%%%%%%%%
%%%%%%%%%%%%%%%%%%%%%%%%%%%%%%%%%%%%%%%%%%%%%%%%%%%%%%%%%%%%%%%%%%%%%%%%%%%%%%%%%%%%%%%%%%%%%%%%%%%%%%
\subsection{Questions}\label{questions}
%%%%%%%%%%%%%%%%%%%%%%%%%%%%%%%%%%%%%%%%%%%%%%%%%%%%%%%%%%%%%%%%%%%%%%%%%%%%%%%%%%%%%%%%%%%%%%%%%%%%%%
%%%%%%%%%%%%%%%%%%%%%%%%%%%%%%%%%%%%%%%%%%%%%%%%%%%%%%%%%%%%%%%%%%%%%%%%%%%%%%%%%%%%%%%%%%%%%%%%%%%%%%

WISeR uses the question tag \texttt{WISeR-question} to denote questions. For yes/no questions, \texttt{WISeR-question} is used in conjunction with the \texttt{polarity} relation to show that the truth value is in question.
\begin{amr}{Did the boy eat lunch?}
(e / eat \\
\hphantom{(e }:actor (b / boy) \\
\hphantom{(e }:theme (l / lunch) \\
\hphantom{(e }:polarity (w / WISeR-question))
\end{amr} \noindent
\begin{amr}{Does the teacher read a lot?}
(r / read \\
\hphantom{(r }:actor (p / person \\
\hphantom{(r :actor (p }:actor-of (t / teach)) \\
\hphantom{(r }:frequency (f / frequent) \\
\hphantom{(r }:polarity (w / WISeR-question))
\end{amr} \noindent
For \textit{wh}-questions such as those containing \textit{who}, \textit{what}, \textit{when}, \textit{where}, \textit{why}, and \textit{how}, \texttt{WISeR-question} is used in the \textit{wh}-item's argument position (e.g., \textit{the boy ate what?}).
\begin{amr}{What did the boy eat?}
(e / eat \\
\hphantom{(e }:actor (b / boy) \\
\hphantom{(e }:theme (w / WISeR-question))
\end{amr} \noindent
\begin{amr}{How fast did the athlete run?}
(r / run \\
\hphantom{(r }:actor (a / athlete) \\
\hphantom{(r }:manner (f / fast \\
\hphantom{(r :manner (f }:degree (w / WISeR-question)))
\end{amr} \noindent
\begin{amr}{Whose toy did the girl find?}
(f / find \\
\hphantom{(f }:actor (g / girl) \\
\hphantom{(f }:theme (t / toy \\
\hphantom{(f :theme (t }:poss (w / WISeR-question)))
\end{amr} \noindent
\begin{amr}{Why did the baby cry?}
(c / cry \\
\hphantom{(c }:actor (b / baby) \\
\hphantom{(c }:cause (w / WISeR-question))
\end{amr} \noindent

For choice questions, we use \texttt{WISeR-choice} to denote options.
\begin{amr}{Do you want tea or coffee?}
(w / want \\
\hphantom{(w }:actor (y / you) \\
\hphantom{(w }:theme (w2 / WISeR-choice \\
\hphantom{(w :theme (w2 }:op1 (t / tea) \\
\hphantom{(w :theme (w2 }:op2 (c / coffee)))
\end{amr} \noindent
\begin{amr}{Did the teacher walk or did she drive to school?}
(s / school \\
\hphantom{(s }:end-of (w / WISeR-choice \\
\hphantom{(s :end-of (w }:op1 (w2 / walk \\
\hphantom{(s :end-of (w :op1 (w2 }:actor (g / girl)) \\
\hphantom{(s :end-of (w }:op2 (d / drive \\
\hphantom{(s :end-of (w :op2 (d }:actor t))))
\end{amr} \noindent
\begin{amr}{Did the man win or lose the lottery?}
(m / man \\
\hphantom{(m }:actor-of (w / WISeR-choice \\
\hphantom{(m :actor-of (w }:op1 (w2 / win \\
\hphantom{(m :actor-of (w :op1 (w2 }:theme (l / lottery)) \\
\hphantom{(m :actor-of (w }:op2 (l2 / lose \\
\hphantom{(m :actor-of (w :op2 (l2 }:theme l))) \\
\end{amr} \noindent
%

%%%%%%%%%%%%%%%%%%%%%%%%%%%%%%%%%%%%%%%%%%%%%%%%%%%%%%%%%%%%%%%%%%%%%%%%%%%%%%%%%%%%%%%%%%%%%%%%%%%%%%
%%%%%%%%%%%%%%%%%%%%%%%%%%%%%%%%%%%%%%%%%%%%%%%%%%%%%%%%%%%%%%%%%%%%%%%%%%%%%%%%%%%%%%%%%%%%%%%%%%%%%%
\subsection{Relative Clauses}\label{relative clauses}
%%%%%%%%%%%%%%%%%%%%%%%%%%%%%%%%%%%%%%%%%%%%%%%%%%%%%%%%%%%%%%%%%%%%%%%%%%%%%%%%%%%%%%%%%%%%%%%%%%%%%%
%%%%%%%%%%%%%%%%%%%%%%%%%%%%%%%%%%%%%%%%%%%%%%%%%%%%%%%%%%%%%%%%%%%%%%%%%%%%%%%%%%%%%%%%%%%%%%%%%%%%%%
Relative clauses are represented with inverse roles. 
\begin{amr}{The boy who wore red sang at the concert}
(s / sing \\
\hphantom{(s }:actor (b / boy \\
\hphantom{(s :actor (b }:actor-of (w / wear \\
\hphantom{(s :actor (b :actor-of (w }:theme (r / red))) \\
\hphantom{(s }:location (c / concert))
\end{amr} \noindent
The main predicate in this sentence is \textit{sing} which therefore forms the root of our annotation. The predicate \textit{wear red} is then introduced with the inverse relation \texttt{actor-of}.
\begin{amr}{The man saw the executive that moved into the large office}
(s / see \\
\hphantom{(s }:actor (m / man) \\
\hphantom{(s }:theme (e / executive \\
\hphantom{(s :theme (e }:actor-of (m / move \\
\hphantom{(s :actor (e :actor-of (m }:end (o / office \\
\hphantom{(s :actor (e :actor-of (m :end (o }:mod (l / large))))
\end{amr} \noindent
Note that the information about the executive moving into a large office is used to identify the person that the man hates. In this sentence, the man saw the executive. In contrast, the following sentence does not involve a relative clause.
\begin{amr}{The man saw that the executive that moved into the large office}
(s / see \\
\hphantom{(s }:actor (m / man) \\
\hphantom{(s }:theme (m / move \\
\hphantom{(s :theme (m }:actor (e / executive) \\
\hphantom{(s :theme (m }:end (o / office \\
\hphantom{(s theme (m :end (o }:mod (l / large))))
\end{amr} \noindent
For this sentence to be true, the man need not directly see the executive. It is sufficient that he sees evidence that the executive is the new occupant of the large office.

%%%%%%%%%%%%%%%%%%%%%%%%%%%%%%%%%%%%%%%%%%%%%%%%%%%%%%%%%%%%%%%%%%%%%%%%%%%%%%%%%%%%%%%%%%%%%%%%%%%%%%
%%%%%%%%%%%%%%%%%%%%%%%%%%%%%%%%%%%%%%%%%%%%%%%%%%%%%%%%%%%%%%%%%%%%%%%%%%%%%%%%%%%%%%%%%%%%%%%%%%%%%%
\subsection{Have-rel-role and have-org-role}\label{have_role}
%%%%%%%%%%%%%%%%%%%%%%%%%%%%%%%%%%%%%%%%%%%%%%%%%%%%%%%%%%%%%%%%%%%%%%%%%%%%%%%%%%%%%%%%%%%%%%%%%%%%%%
%%%%%%%%%%%%%%%%%%%%%%%%%%%%%%%%%%%%%%%%%%%%%%%%%%%%%%%%%%%%%%%%%%%%%%%%%%%%%%%%%%%%%%%%%%%%%%%%%%%%%%

WISeR follows AMR in using special predicate to attribute certain roles to people. For instance, a person who stand in a certain professional or personal relation to another.
\begin{amr}{she is my doctor}
(h / have-rel-role \\
\hphantom{(h }:actor (s / she) \\
\hphantom{(h }:theme (i / i) \\
\hphantom{(h }:attribute (d / doctor))
\end{amr} \noindent
\begin{table}[!h]
    \centering
    \begin{tabular}{| c || c |}
        \hline
        \texttt{actor} & person who has role \\ \hline
        \texttt{theme} & with whom \\ \hline
        \texttt{attribute} & the relation \\  \hline
    \end{tabular}
    \caption{List of arguments for \texttt{have-rel-role}}
    \label{tab:have-rel}
\end{table}
\begin{amr}{My girlfriend swims}
(s / swim \\
\hphantom{(s }:actor (p / person \\
\hphantom{(s :actor (p }:actor-of (h / have-rel-role \\
\hphantom{(s :actor (p :actor-of (h }:theme (i / i) \\
\hphantom{(s :actor (p :actor-of (h }:attribute (g / girlfriend))))
\end{amr} \noindent
Other examples of \texttt{have-rel-role} include: \textit{father, sister, husband, grandson, godfather, stepdaughter, brother-in-law, friend, boyfriend, buddy, enemy, landlord, tenant} etc.

We use a similar structure for \texttt{have-org-role}.
\begin{table}[!h]
    \centering
    \begin{tabular}{| c || c |}
        \hline
        \texttt{actor} & person who has role \\ \hline
        \texttt{theme} & organization \\ \hline
        \texttt{attribute} & the role \\  \hline
    \end{tabular}
    \caption{List of arguments for \texttt{have-org-role}}
    \label{tab:have-org}
\end{table}
\begin{amr}{She is the company president}
(h / have-org-role \\
\hphantom{(h }:actor (s / she) \\
\hphantom{(h }:theme (c / company) \\
\hphantom{(h }:attribute (p / president))
\end{amr} \noindent

\end{document}